\documentclass[acmtog]{acmart}
\acmSubmissionID{346}

\usepackage{booktabs} % For formal tables
\usepackage{makecell}
\usepackage{color}
\usepackage{graphicx}
\usepackage{subfigure}
\usepackage{float}
\usepackage{xcolor}
\usepackage{multirow}
\usepackage{bm}

\usepackage{amsfonts,amssymb,amsmath,amsthm}
\usepackage{bbding}
\usepackage{pifont}
\usepackage{wasysym}
\usepackage{lineno}
\usepackage[normalem]{ulem}

\usepackage[ruled]{algorithm2e} % For algorithms

\SetAlFnt{\small}
\SetAlCapFnt{\small}
\SetAlCapNameFnt{\small}
\SetAlCapHSkip{0pt}

\copyrightyear{2022}
\acmYear{2022}
\setcopyright{acmcopyright}
\acmConference[SA '22 Conference Papers]{SIGGRAPH Asia 2022 Conference Papers}{December 6--9, 2022}{Daegu, Republic of Korea} 
\acmBooktitle{SIGGRAPH Asia 2022 Conference Papers (SA '22 Conference Papers), December 6--9, 2022, Daegu, Republic of Korea}
\acmPrice{15.00}
\acmDOI{10.1145/3550469.3555399}
\acmISBN{978-1-4503-9470-3/22/12}

\begin{document}
% Title portion
\title{VideoReTalking: Audio-based Lip Synchronization for Talking Head Video Editing In the Wild}

\author{Kun Cheng}
\authornote{Both authors contributed equally to this research. Xiaodong Cun is the corresponding author. Project page: https://vinthony.github.io/video-retalking/ }
\affiliation{
    \institution{Xidian University}
    \city{Xi’an}
    \country{China}
    }
\email{kuncheng.xidian@gmail.com}

\author{Xiaodong Cun}
\authornotemark[1]
\affiliation{
    \institution{Tencent AI Lab}
    \city{Shenzhen}
    \country{China}
    }
\email{vinthony@gmail.com}

\author{Yong Zhang}
\affiliation{
    \institution{Tencent AI Lab}
    \city{Shenzhen}
    \country{China}
     }
\email{zhangyong201303@gmail.com}

\author{Menghan Xia}
\affiliation{
    \institution{Tencent AI Lab}
    \city{Shenzhen}
    \country{China}
     }
\email{menghanxyz@gmail.com}

\author{Fei Yin}
\affiliation{
    \institution{Tsinghua University}
    \city{Shenzhen}
    \country{China}
    }
\email{yinf20@mails.tsinghua.edu.cn}

\author{Mingrui Zhu}
\affiliation{
    \institution{Xidian University}
    \city{Xi'an}
    \country{China}
    }
\email{mrzhu@xidian.edu.cn}

\author{Xuan Wang}
\affiliation{
    \institution{Tencent AI Lab}
    \city{Shenzhen}
    \country{China}
     }
\email{xwang.cv@gmail.com}

\author{Jue Wang}
\affiliation{
    \institution{Tencent AI Lab}
    \city{Shenzhen}
    \country{China}
    }
\email{arphid@gmail.com}

\author{Nannan Wang}
\affiliation{
    \institution{Xidian University}
    \city{Xi'an}
    \country{China}
    }
\email{nnwang@xidian.edu.cn}

% \sout{\cite[{p. 10}]{lamport}}

% \author{Kun Cheng$^{1,2*}$, Xiaodong Cun$^{2*}$, Yong Zhang$^{2}$, Menghan Xia$^{2}$, Xuan Wang$^{2}$, Fei Yin$^{2,3}$, Mingrui~Zhu$^{1}$, Jue Wang$^{2}$, Nannan Wang$^{1}$}
% \affiliation{%
%   \institution{$^{1}$State Key Laboratory of Integrated Services Networks, Xidian University, China\\
%   $^{2}$Tencent AI Lab, China\\
%   $^{3}$Tsinghua Shenzhen International Graduate School, Tsinghua University, China\\
%   }
% }
% \thanks{*Both authors contributed equally to this research.}
% \email{kuncheng.xidian@gmail.com, vinthony@gmail.com}
% \url{http://vinthony.github.io/video-retalking}

\renewcommand\shortauthors{Cheng and Cun, et al.}

\begin{abstract}
% 1. lipsync/video editing current challenge [single person method, low-resolution & lack of accurate lipsync.]
% 2. we propose a framework which can
%   - high-resolution generation
%   - accurate lip sync 
%   - editable expression

% Yong 
% 1. Goal of video dubbing
% 2. summary of current challenges or drawbacks 
%   - most are personalized 
%   - few focus on generic 
%   - low quality: low resolution, blur, teeth, inconsistency 
% 3. work summary
%   - a novel framework for generic video dubbing 
%   - High resolution and high-fidelity and details 
%   - enable emotion control 
% 4. How to do 
%   - "removing-then-editing"
%   - a module for emotion removing
%       - stability of audio driving, simplify the input
%   - a module for editing 
%       - audio injection 
%       - emotion control 
%   - a module for quality 
%       - id preserving 
%       - facial details:  teeth, eyes 
%  5. experiments for validation

We present VideoReTalking, a new system to edit the faces of a real-world talking head video according to input audio, producing a high-quality and lip-syncing output video even with a different emotion.
%~(even with different emotions) under the new audio. 
% Generally, our system fellows the idea of removing the talking-related motion from the original video, and then, generating the audio-driven lips from the new face. 
Our system disentangles this objective into three sequential tasks: (1) face video generation with a canonical expression; (2) audio-driven lip-sync; and (3) face enhancement for improving photo-realism. 
%To achieve this goal, we propose three networks for expression reenactment, lip-sync, and face enhancement.
%Given a talking-head video, we first edit all the expression of the original video frame-wisely by the same expression template using an expression editing network, causing a video with the frozen expression for lip synthesis.
Given a talking-head video, we first modify the expression of each frame according to the same expression template using the expression editing network, resulting in a video with the canonical expression. This video, together with the given audio, is then fed into the lip-sync network to generate a lip-syncing video.
% including a module to stabilize the expression in the original videos by a video self-decomposed network, which results will be edited by a fixed expression parameters to produce the basic emotion of the audio. 
% And then, the edited face will be served as a stable reference for lip sync network, and it will synthesis the motion of the lower half face.
Finally, we improve the photo-realism of the synthesized faces through an identity-aware face enhancement network and post-processing.
We use learning-based approaches for all three steps and all our modules can be tackled in a sequential pipeline without any user intervention.
% Furthermore, our system is a generic approach that \red{does not need to be} retrained to a specific video or person. 
Furthermore, our system is a generic approach that does not need to be retrained to a specific person. 
Evaluations on two widely-used datasets and in-the-wild examples demonstrate the superiority of our framework over other state-of-the-art methods in terms of lip-sync accuracy and visual quality.
% The proposed framework have two major differences between the proposed method and the previous arts. On the one hand, we propose a more general solution which can works on any identify and the real-world videos. On the other hand, our method can edit the emotion~(\emph{e.g.} smile and neutral) and the lip-sync concurrently.
% Unlike previous works which only generate low-quality videos or for personalized video portrait, our method synthesize high-quality results for any real-world talking videos. 
% In detail, to control the expression and generate the stable results of the lip sync network, we first learn a self-decomposed face reenactment network in a self-supervised manner to decouple the motion and expression of a talking video. Then, we use a lip synthesis network to modify the lip related motions, where our lip synthesis network utilize the edited videos for a stable expression reference. Finally, to enhance the visual quality of the produced results, we leverage a GAN prior network to enhance the synthesized video. Thanks to the provided stable reference, our methods can generate accurate lip synthesis results with the reference emotion unchanged. Experiments show the advantage of the proposed methods.

% Audio-based visual dubbing aims to edit the lips related motions of the talking video by the given audio. Previous methods are either struggling to synthesize the photo-realistic images or focusing on the personalized video portrait personalized video portrait. 
\begin{figure}
    \centering
    \includegraphics[width=1\columnwidth]{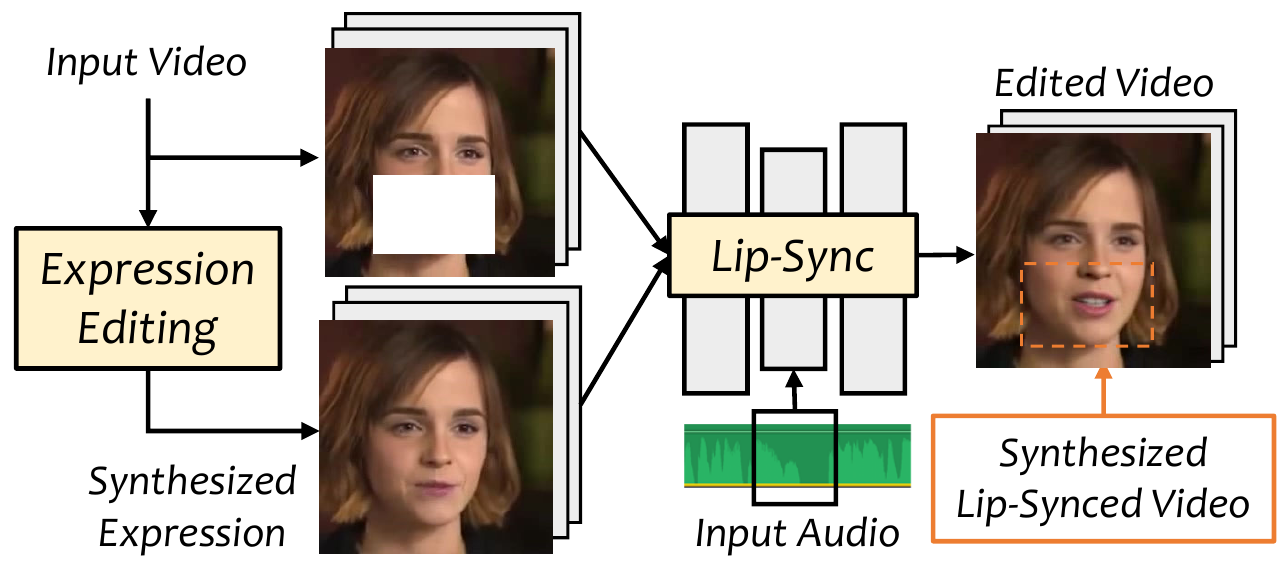}
    % \vspace{-2em}
    \caption{Our method modifies the original video and generates a lip-syncing video by an input audio through expression editing and lip-sync networks. Natural face \copyright~\emph{ONU Brasil}~(CC BY).}
    % \vspace{-1em}
    \label{fig:overview}
\end{figure}
\end{abstract}

\begin{teaserfigure}
\centering
  \includegraphics[width=1\textwidth]{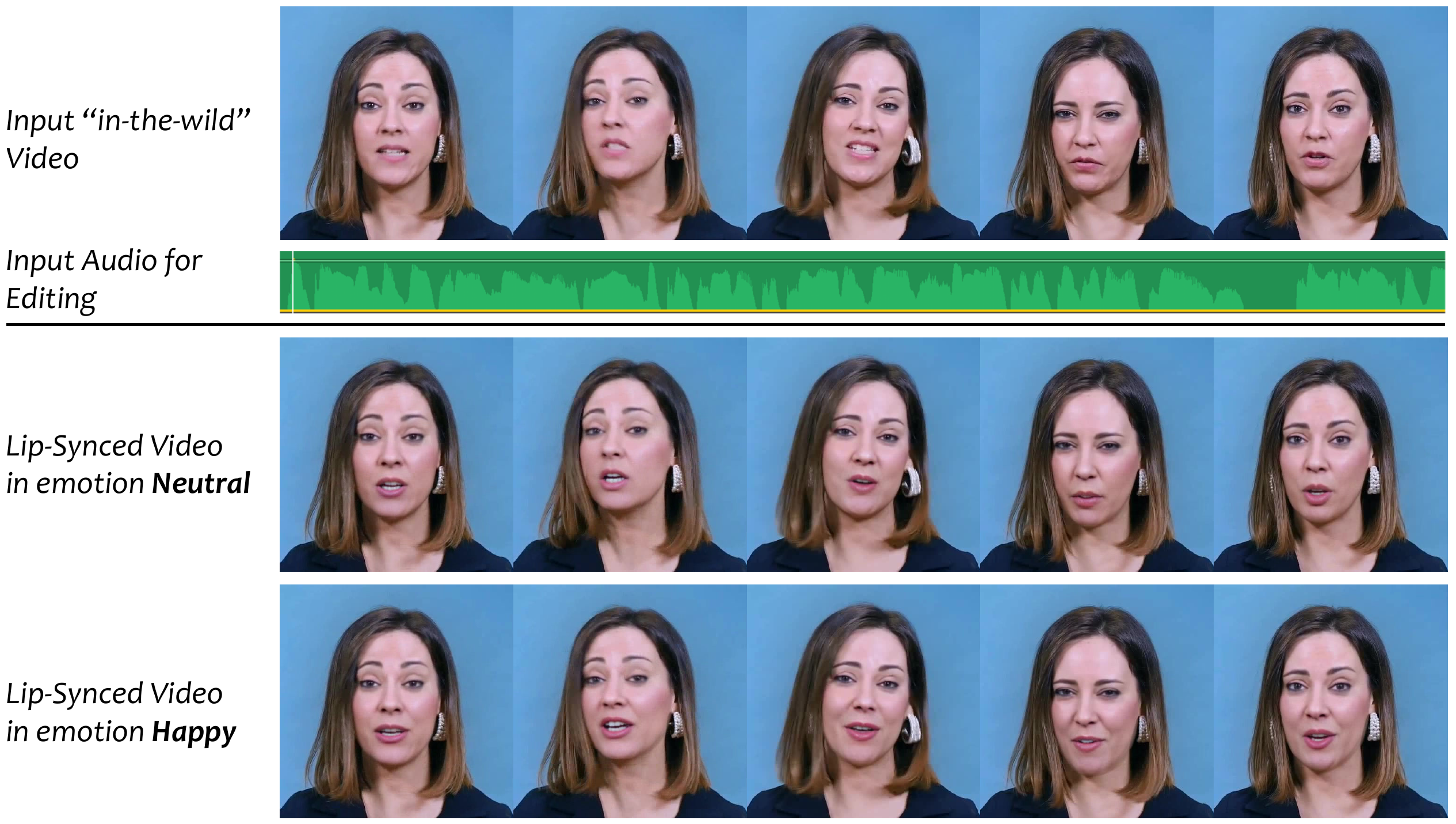}
%   \vspace{-1em}
  \caption{Given an arbitrary talking video and another audio, our method can synthesize a photo-realistic talking video with accurate lip-audio synchronization with retouched face expressions. Natural face \copyright~\emph{European Central Bank}~(CC BY).
  }
  \setlength{\abovecaptionskip}{-0.1cm}   % distance between the caption and Figure
  \label{fig:teaser}
\end{teaserfigure}

%
% The code below should be generated by the tool at
% http://dl.acm.org/ccs.cfm
% Please copy and paste the code instead of the example below.
%
% \begin{CCSXML}
% <ccs2012>
%   <concept>
%       <concept_id>10010147.10010371.10010382.10010385</concept_id>
%       <concept_desc>Computing methodologies~Image-based rendering</concept_desc>
%       <concept_significance>500</concept_significance>
%       </concept>
%  </ccs2012>
% \end{CCSXML}

% \ccsdesc[500]{Computing methodologies~Image-based rendering}


% \ccsdesc[500]{Computing methodologies~Computer vision}
% \ccsdesc[300]{Computing methodologies~Neural networks}

% \keywords{Facial Animation, Neural Networks}

%
% End generated code
%

% \keywords{Wireless sensor networks, media access control,
% multi-channel, radio interference, time synchronization}

% \settopmatter{printacmref=false} % Removes citation information below abstract
% \renewcommand\footnotetextcopyrightpermission[1]{} % removes 

\maketitle

\newcommand{\xiaodong}[1]{{\color{blue}{[xiaodong: #1]}}}
\newcommand{\yong}[1]{{\color{red}{[yong: #1]}}}
\newcommand{\kun}[1]{{\color{green}{[kun: #1]}}}
\newcommand{\menghan}[1]{{\color{orange}{[menghan: #1]}}}

\section{Introduction}
\begin{figure*}[t]
\begin{center}
\centerline{\includegraphics[width=1\linewidth]{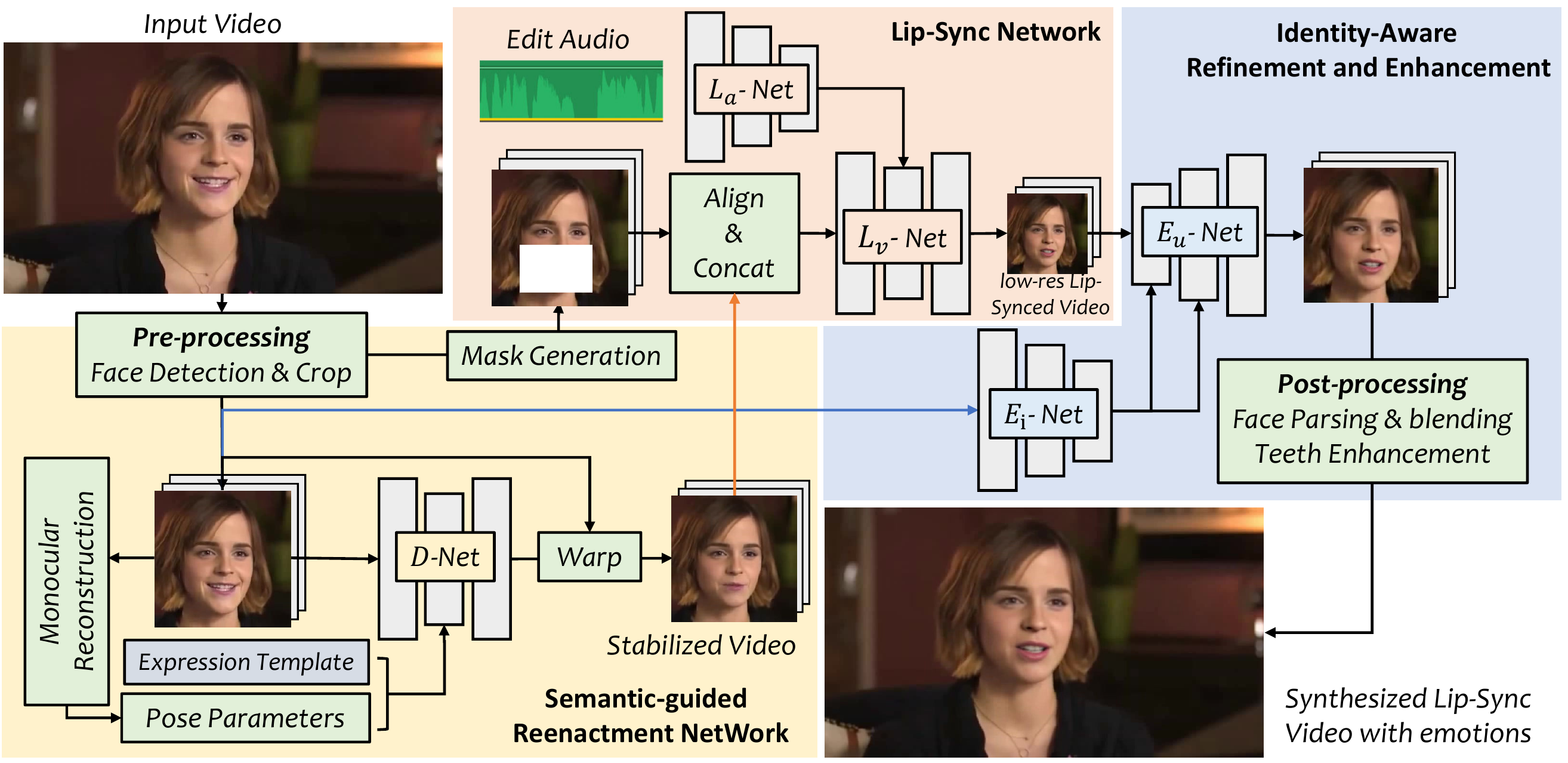}}
% \vspace{-1.5em}
\caption{Our framework contains three main components for photo-realistic lip-sync video generation. Natural face \copyright~\emph{ONU Brasil}~(CC BY).}
\label{fig:pipeline}
\vspace{-2em}
\end{center}
\end{figure*}

The task of editing a talking head video according to an input speech audio has important real-world applications, such as translating an entire video into a different language, or modifying the speech after video recording. Known as visual dubbing, this task has been studied in several prior works~\cite{obama,AudioDVP,NVP,wav2lip}, which edit the input talking-head video by modifying the facial animation and emotion to match the target audio, while leaving all the other motions unchanged~(shown in Figure~\ref{fig:teaser}).
Some methods~\cite{obama,AudioDVP,NVP} can achieve satisfactory results on a specific speaker, but require training on the talking corpus of the target speaker to obtain a personalized model, which is not always available. %So, the fast forward based generic model is more promising.
On the other hand, the current generic methods produce blurry lower faces~\cite{wav2lip} or inaccurate lip synchronization~\cite{song2022everybody}, 
% \menghan{it is better to give an brief explanation about why this limitation exist, e.g., the difficulties of the task or the ignored issues in their methods, etc} 
which are visually intruding. These methods also do not support emotion editing, which is often desirable when changing the speech content.

% \begin{figure}
%     \centering
%     \includegraphics[width=0.9\columnwidth]{figs_pdf/overview_vrt.pdf}
%     % \vspace{-2em}
%     \caption{Our method modifies the original video and generates a lip-syncing video by an input audio through expression editing and lip-sync networks.}
%     % \vspace{-1em}
%     \label{fig:overview}
% \end{figure}

% % 3. Our solution
Inspired by previous inpainting-based talking-head video editing approaches~\cite{wav2lip}, we present a new system to edit the talking lips to match the input audio with more stable lip-sync results and better visual quality. 
Previous works consider the original frames in the video as the head pose references. 
However, we have found that lip generation is very sensitive to these references, and directly using original frames as basis for lip generation often produces out-of-sync results. 
To this end, as shown in Figure~\ref{fig:overview}, we employ a divide-and-conquer strategy by neutralizing the facial expressions first, then use the modified frames as pose references for lip generation, which is more accurate given that all reference faces now have the same canonical expression. %we edit the expression of the original video firstly, and then, the new month shape will also be inherited to the lip-sync network, resulting an emotional talking-head generation by changing the expression parameters of the references.
% and synthesizing the talking-head videos with the stable expression. 
Finally, in contrast to previous works that often produce low-resolution and blurry results, we produce photo-realistic results via the proposed identity-aware enhancement network and the restorations ~\cite{gpen,gfpgan} based on StyleGAN's facial prior~\cite{stylegan}. 
%StyleGAN~\cite{stylegan} prior restorations~\cite{gpen,gfpgan}.

Specifically, given an arbitrary talking video, we first crop the face region and extract the pose and expression coefficients of the 3D Morphable Model~(3DMM) by a deep neural network~\cite{face3d}. 
We then use the parameters of the 3DMM with a standard neutral template expression
%a standard neutral template expression parameters of the 3DMM 
and re-generate a video through the semantic-guided expression reenactment network similar to \cite{pirender}. 
By doing so, we obtain a video with the same canonical expression across all the frames, and they will be considered as the structure references for our lip-sync network. 
Interestingly, we can also synthesize talking head videos with different emotions by changing the expression template. 
For example, by changing the lip shape of the expression template to match the ``happy" emotion, this lip shape will be taken into account in the lip-sync network, causing the talking-head video exhibits the same emotion.

After expression neutralization, a lip-sync network is then applied to synthesize photo-realistic lower-half faces using the synthesized expression as the conditional structure information.
Specifically, we design an hourglass-like network with the Fast Fourier Convolution block~\cite{ffc} as the basic learning unit, since it achieves great success in the general image inpainting task~\cite{lama}. 
As for the audio injection, we use the Adaptive Instance Normalization~(AdaIN) block~\cite{huang2017adain} to modulate the audio features in global. 
Similar to \cite{wav2lip}, we use a pre-trained lip-sync discriminator to ensure the audio-visual synchronicity. 

Although previous steps can synthesize talking-head videos with relatively accurate lip shapes, the visual quality is still limited by the low-resolution training  datasets~\cite{voxceleb,lrs2}. 
To solve this problem, we design an identity-preserving face enhancement network to produce high-quality outputs by  progressive training. 
The enhancement network is trained on an enhanced LRS2 dataset~\cite{lrs2} enhanced by the face restoration method~\cite{gpen}. We also apply the StyleGAN prior guided face restoration network~\cite{gfpgan} to remove visual artifacts around the teeth.

All the above modules can be applied in sequential order without manual intervention or fine-tuning. 
We conduct extensive experiments to evaluate our framework on several existing benchmarks as well as in-the-wild videos.  
Results show that the proposed system can produce videos with much higher visual quality than previous methods while providing accurate lip synchronization.

% % % 4. contribution
% We summary our contributions as fellow:

% \begin{itemize}
%     \item We propose a new method to edit the talking-head video in the wild.
%     \item Our method achieves the state-of-the-art performance with respect to the video quality and lip-sync results.
% \end{itemize}

% % \clearpage

\section{Related Work}
We review the related methods from two aspects, including the visual dubbing task which aims to edit the input video through audio, and single image animation using the audio as conditions.
% into two categories based on the type of input visual information:  video-to-video methods and one-shot methods. The visual input of the former methods is a video, and the visual input of the latter methods is a single reference image. 

\subsection{Audio-based Dubbing in Video Editing}

\subsubsection{Arbitrary-subject methods}
% 1. Arbitrary
% Speech2Vid; LipGAN,Wav2Lip; Everybody’s Talkin’; MM22; SyncFaceTalk; 
Arbitrary-subject methods aim at building a general model that does not need to be retrained for different identities.
Speech2Vid~\cite{Chung17b} can re-dub a source video with a different segment of audio thanks to the context encoder. %which help reproduce the background. 
%The generated face can be pasted to original video. 
Reconstructing the lower-half face by inpainting is popular recently~\cite{lipgan, wav2lip, park2022synctalkface}. For example, LipGAN~\cite{lipgan} design a neural network to fill the lower-half face as a pose prior. Wav2Lip~\cite{wav2lip} extends LipGAN using a pre-trained SyncNet as the lip-sync discriminator~\cite{syncnet} to generate accurate lip synchronization. Based on Wav2Lip, SyncTalkFace~\cite{park2022synctalkface} involve the audio-lip memory to store the lip motion features implicitly and retrieve them at inference time. Another category of the methods predicts the intermediate representation first, and then, synthesizes the photo-realistic results by image-to-image translation networks, for example, the facial landmarks~\cite{xie2021towards} and the facial landmarks based on 3D face reconstruction~\cite{song2022everybody}. However, all these methods are struggling to synthesize the high-quality results with editable emotion. 

%Some methods do not reconstruct lip region directly but generate latent representations firstly from audio and then render them to video. 
% propose a two-stage paradigm employs facial landmarks as latent representations to predict lip movements. \cite{song2022everybody} generates 3DMM expression coefficients and use a arbitrary-subject video rendering network to construct photo-realistic video.
%But all the above methods can only handle low-resolution videos. It is difficult to collect a large-scale high-resolution talking face dataset. And it is also really challenging to learn robust audio-visual correlation on small-scale HD dataset. Our method extends the audio-visual correlation learned on low-resolution dataset to high-resolution dataset rather than directly learning on it which cleverly avoids this problem. To the best of our knowledge, our proposed system is the first arbitrary-subject audio-driven visual dubbing method that can handle high-resolution videos.

\subsubsection{Personalized methods}
% 2. One person one model
% Audio-driven: SyncObama; Neural Voice Puppetry; Live Speech Portraits;  LipSync3d; AD-Nerf; FACIAL; EVP
% Video-driven: DVP; Neural-style preserved;
% text-driven: write-a-speaker;

% landmark-based: EVP; Live Speech Portraits
% 3d mesh: LipSync3D
% 3dmm: NVP; FACIAL
% nerf: AD-Nerf
%Personalized Visual Dubbing is concerned with generating synchronizing lip motions with arbitrary speech input for one specific target. 
Personalized visual dubbing is easier than the generic one, since these methods are limited to the certain person in the known environment.
For example, SynthesizeObama~\cite{obama} can synthesize the mouth region of Obama by the audio-to-landmark network. Inspired by the face reenactment methods~\cite{dvp, thies2019deferred}, recent visual dubbing methods focus on generating the intermediate representation from audio, and then, rendering the photo-realistic results by the image-to-image translation networks. For example, several works~{\cite{NVP, AudioDVP,zhang2021facial}} focus on the expression coefficient from the audio features and render the photo-realistic results by the image generation networks~\cite{thies2019deferred, dvp, vid2vid}. Facial landmarks~\cite{lu2021live} and edges~\cite{EVP} are also popular choices by projecting the 3D rendered faces since it contains sparser information.
Furthermore, 3D mesh-based~\cite{lipsync3d} and NeRF~\cite{nerf}-based methods~\cite{adnerf} are also powerful.
% \textcolor{red}{LipSync3D~\cite{lipsync3d} normalizes the input face on pose and lighting, and then utilizes 3D mesh as the intermediate representation to generate video.} 
% \textcolor{red}{NeRF~\cite{nerf}-based method~\cite{adnerf} feeds audio into a conditional implicit function to generate a dynamic neural radiance field.}
% Inspired by recently development of Neural Radiance Fields~(NeRF), audio can also be represented as the scene fields. 
Although these methods can synthesize the photo-realistic results, they have relatively limited applications because they need to retrain the model on the specific person and environment.

\nocite{wang2021hfgi}

\subsection{Audio-based Single Image Facial Animation}
% one-shot
% X2face; ATVG; DAVS; RNN; Speech-Driven Animation; Flow-guided; PC-AVS; MakeItTalk; Audio2Head,AAAI-22; ijcai22; StyleHEAT
Different from the visual dubbing, single image face animation aims to generate the animation by single driven audio, and it has also been influenced by the video-driven face animation. For example, \cite{song2018talking} generate the motion from audio using the recurrent neural network, \cite{zhou2019talking} disentangle the input to subject-related information and speech-related information by adversarial representation learning. \cite{vougioukas2020realistic, pc-avs} consider the audio as the latent code and drive the face animation by an image generator.
% propose a temporal GAN and three different discriminators to generate lip movements and facial expressions. PC-AVS~\cite{pc-avs} utilize a style-based generator to reconstruct face and the audio servers as the latent code.
The intermediate representation is also a popular choice in this task. ATVG~\cite{chen2019hierarchical} and MakeItTalk~\cite{makeittalk} first generate the facial landmarks from audio, and then, render the video using a landmark-to-video network. Dense flow field is another active research direction~\cite{2203.04036, fomm}. \cite{hdtf} predict the 3DMM coefficients from audio and then transfer these parameters into a flow-based warping network. \cite{wang2021one, wang2021audio2head} borrow the idea from video driven face animation~\cite{fomm}.

\section{Framework}
Technically, our method is a cross-modal video inpainting framework to fill the masked lower-half face under the guidance of the driven audio and the emotion-modulated reference frame. 
To this end, we design a lip-sync network~($L$-Net in Sec.~\ref{sec:lipsyncnet}), which uses the masked lower-half face frames, the given audio, and the original video frames as input to generate a lip-syncing video. 
However, there are two major problems if we use the $L$-Net only. The first is the information leak caused by the reference frame, where the generated lip still relies on the reference heavily. The other is the low visual quality since current large-scale talking head datasets are in low resolution.

To this end, except $L$-Net, we propose two additional modules as shown in Figure~\ref{fig:pipeline}. 
First, to solve the information leak, we generate a video with the frozen face expression by a semantic-guided expression reenactment network~(${D}$-Net in Sec.~\ref{sec:pirender}). 
The synthesized lips are the reference lips instead of the original ones. 
Then, the lower-half faces of the edited video will be used as a reference structure for our lip-synthesis network~(${L}$-Net). In ${L}$-Net, our method takes the audio as input and synthesizes the lip-sync results frame-wisely. Furthermore, we design an ${E}$-Net for the identity-aware face restoration in Sec.~\ref{sec:enhance}. Finally, we can paste the generated face back to the original video seamlessly through the post-processing in Sec.~\ref{sec:post}. Below, we give the details of each component.

\begin{figure}
    \centering
    \includegraphics[width=1\columnwidth]{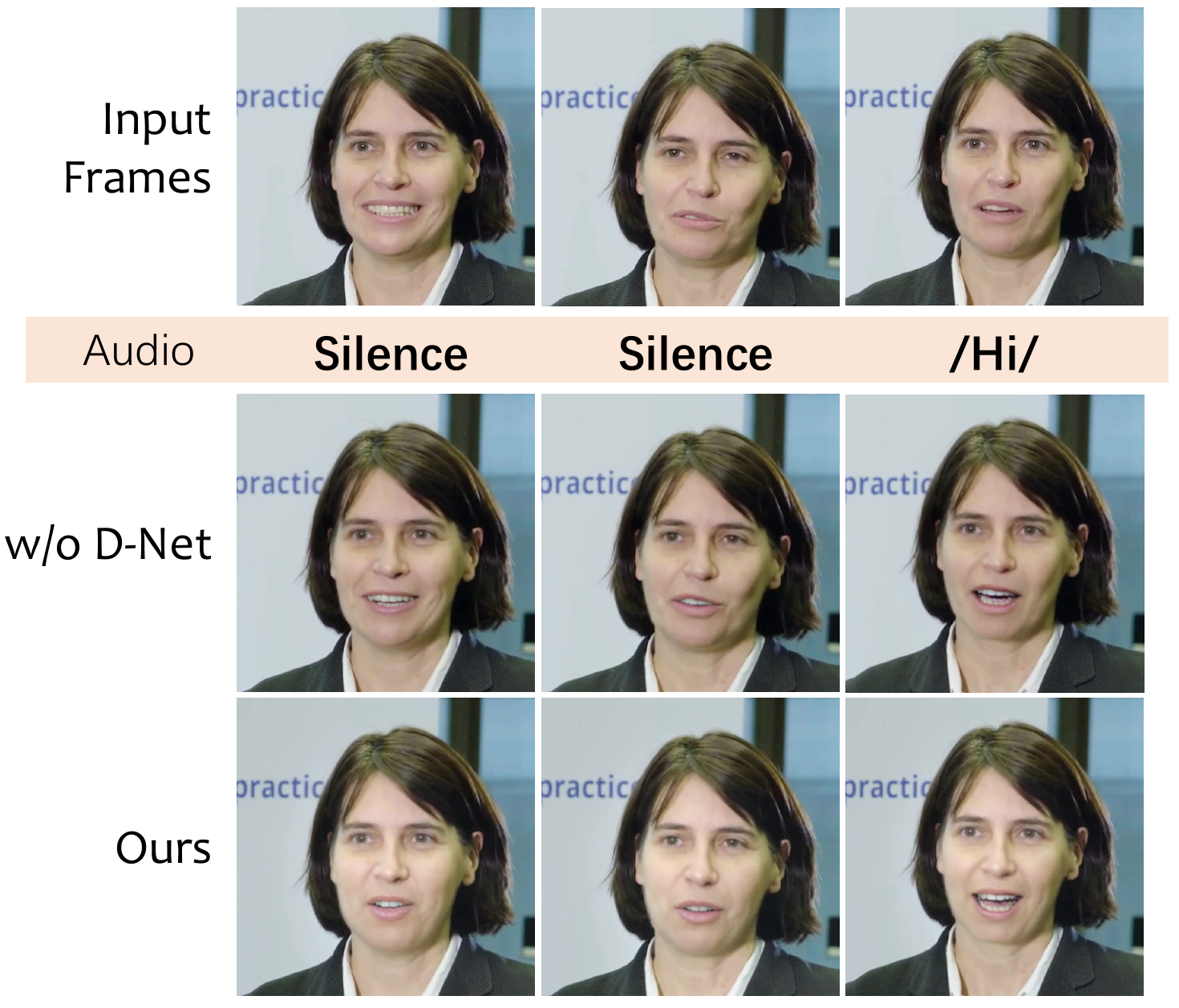}
    % \vspace{-2.5em}
    \caption{ 
    % \sout{The proposed D-Net is used to remove the talking-related motions from the original video. Otherwise, the generated month shape will be hugely influenced by the original mouth. We show the final results of our framework when the audio is \texttt{silence} and say \textit{/Hi/}, where D-Net can successful prevent the lip shape leak.} 
    The proposed D-Net is used to remove the talking-related motions from the original video. W/o D-Net, the generated lip motion is heavily influenced by the source video and is still moving even when the audio is silent, indicating that information leakage affects lip synthesis. Natural face \copyright~\emph{European Central Bank}~(CC BY).
    % The proposed D-Net can remove the talking-related motions from the original video for more accurate lip synchronization.
    }
    % \vspace{-4mm}
    \label{fig:dnet}
\end{figure}

\subsection{Semantic-guided Reenactment Network}
\label{sec:pirender}
% \xiaodong{add the motivation of more stable results}
% One challenge of current audio-based talking head video editing~\cite{wav2lip} is the information leak by the reference face in the test time. To solve this problem, 
% Editing the lip-related motion in \red{the} video directly is challenging, previous works often omit the original lip motion changes~\cite{wav2lip} or retiming the background~\cite{obama,song2022everybody} to avoid unnatural movements between the head pose and lip.
It is challenging to edit the lip-related motion in the video directly.
Previous works often omit the original lip motion changes~\cite{wav2lip} or retiming the background~\cite{obama,song2022everybody} to avoid unnatural movements between the head pose and lip.
Differently, we directly edit the whole lower-half face, including the facial movements with the help of a face reenactment method. 
Our key observation is that there is an information leak~\cite{lipgan,wav2lip} in conditional in-painting based methods if we use the original frame as the conditional image for lip synchronization. 
We give an example to show this phenomenon in Figure~\ref{fig:dnet}. 
Given the audio and the input frames, if we directly use the original frames as reference~(w/o $D$-Net), the generated lips will be modified according to the original one. 
Thus, we aim at editing the expression of the whole lower-half face by the proposed semantic-guided reenactment network. 
Then, the frame with stable expression will serve as the reference for further lip synthesis. 

% Different from previous works which use the original face~\cite{wav2lip} or background retiming~\cite{obama,song2022everybody}, we edit the expressions of the whole lower-half face by the face reenactment network with semantic coefficient guidance.
As shown in Figure~\ref{fig:pipeline}, after the face detection and crop, we extract the pose and expression coefficients from each frame using monocular face reconstruction~\cite{face3d}. 
Then, we obtain the new driven signal by replacing the original expression coefficient with the pre-defined expression template. 
Thus, we can synthesize a video with the frozen expression via the produced dense warp fields of the network and the original frame. 
Similar to \cite{pirender}, the ${D}$-Net contains two encoder-decoder-like structures for coarse-to-fine training. After the expression editing, we get the stabilized expression across all the frames. Note that, since the quality of the face reenactment network is still limited, we use the edited face as the structure reference of our lip-sync network. To this end, we first detect facial landmarks, smooth them utilizing a temporal Savitzky–Golay filter, and then use the keypoints of the eye center and the nose as anchors for face alignment.

% The difference between source frame and the driving frame in the original video as an identity frame and drive the other frames using the extracted 3DMM parameters. 
%Then, we align the faces of the original frame and the generated face by the facial key-points based 
Interestingly, we can also utilize this information leak caused by the lip-sync reference frame through more expression templates~(\emph{e.g.} smile), resulting in an emotional talking-face video as shown in Figure~\ref{fig:teaser}. Since our expression reenactment network only edits the lower-half face of the original video, inspired by the facial action code system~\cite{facs}, we can generate the talking faces in other emotions, \emph{i.e.}, anger and surprise, via the image-based expression editing network~\cite{ganimation} on the upper face.  We consider it as a plugin and show some results in Sec.\ref{sec:result}.

% \xiaodong{\\
% In semantic-guided video self-decomposition, add a figure about alignment: \\
% 1. comparison between different face alignment method by visualize the movement of landmarks with the ground truth face edge(0-17 landmarks)] \\
% 2. add a figure about the effect of D-Net.
% }
% \begin{figure}
%     \centering
%     \includegraphics[width=\columnwidth]{figs_pdf/align.pdf}
%     \caption{Facial landmark tracemaps of different alignment methods. (a)use frame crop. (b)5 landmarks. (c)stit. (d)GT }
%     \label{fig:align}
% \end{figure}

% Previous work~\cite{wav2lip} use a random choice reference for talking-face generation. However, due to the One challenge of previous talking face generation network is that 

% \subsubsection{The comparison of different Face Alignment~(maybe)}

\begin{figure}
    \centering
    \includegraphics[width=1\columnwidth]{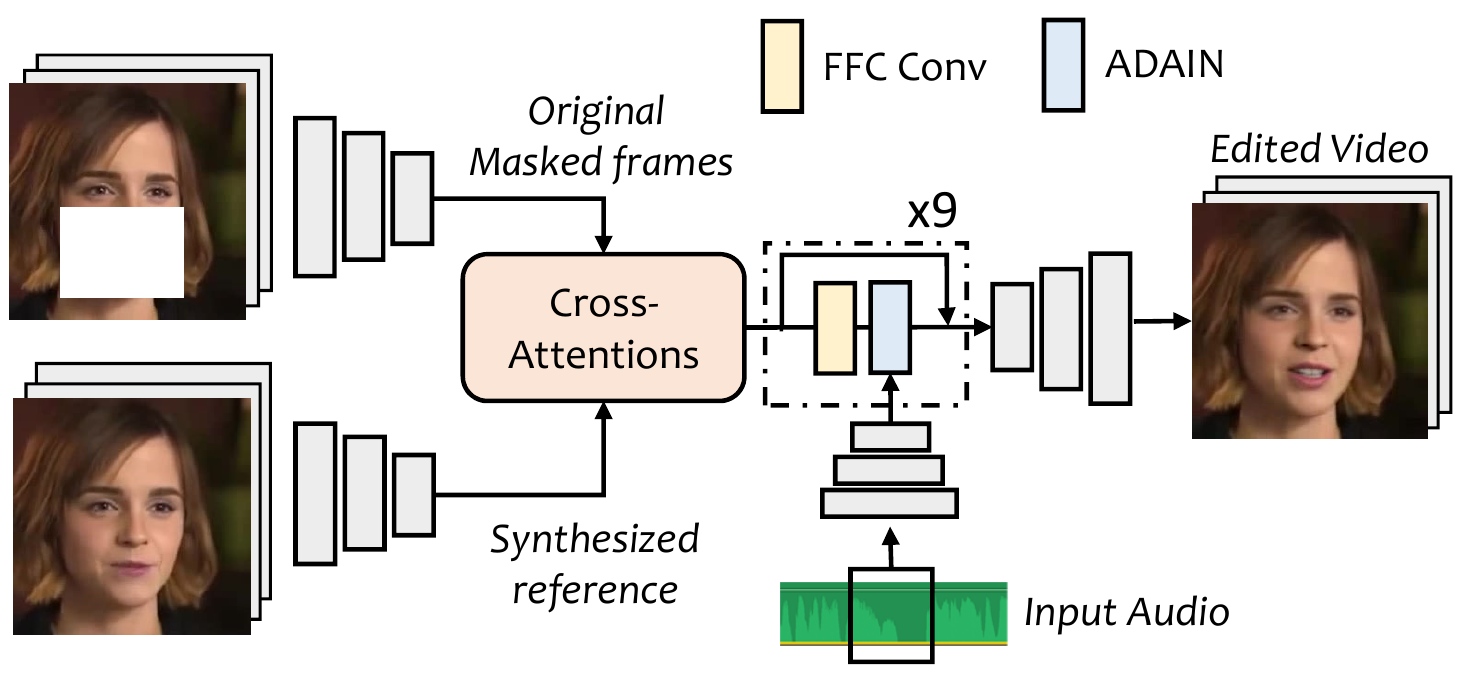}
    % \vspace{-2.5em}
    \caption{The detailed structure of the proposed $L$-Net. The skip-connections between the  reference features and decoder are omitted for clarify. Natural face \copyright~\emph{ONU Brasil}~(CC BY).}
    % \vspace{-2.2em}
    \label{fig:lipnet}
\end{figure}

\subsection{Lip-Sync Network}
\label{sec:lipsyncnet}
Our lip-sync network~($L$-Net) is inspired by a recent conditional inpainting-based framework~\cite{wav2lip}, which edits the original video directly through new audio. 
Differently, we use the pre-processed frames from ${D}$-Net as the identity and structure reference, the audio and the masked original frames as the condition, to synthesize the lip-syncing video with respect to the input audio. 

In Figure~\ref{fig:pipeline}, we give a brief overview of $L$-Net, which contains two sub-networks, ${L}_{a}$ and ${L}_{v}$, for audio and video processing, respectively. 
Here, we give the detailed structure of ${L}$-Net in Figure~\ref{fig:lipnet}.
For the audio processing, we firstly extract the mel-spectrograms from the raw audio and use a ResNet-based encoder~\cite{resnet} to extract the global audio vector $F_{audio}\in\mathbb{R}^{256 \times 1 \times 1}$ of a time window. 
Following previous works, the time window is set to 0.2s per frame, causing the feature in the dimension of 80$\times$16 to process.
As for the image generation, we first extract the image features $F_{ref}, F_{orig}\in\mathbb{R}^{256 \times H \times W}$ from the pre-processed referenced images and the original masked image by two different encoders respectively, then, these features are learned to model the relationship between pixels automatically via two cross-attention blocks~\cite{attention}. These cross-attention blocks will calculate the pixel-wise corresponding matrix of two features and enlarge the reception fields. After that, we use nine residual Fast Fourier Convolutional blocks~\cite{ffc} to refine the features inspired by recent general image inpainting framework~\cite{lama}, and we inject the audio features by the AdaIN blocks~\cite{huang2017adain} which normalize visual features channel-wise after each FFC block. Finally, a series of the convolutional up-sampling layers are used to generate the final results.

% We give more details of the network structure in the supplementary materials.

\subsection{Identity-aware Enhancement Network}
\label{sec:enhance}
\begin{figure}
    \centering
    \includegraphics[width=1\columnwidth]{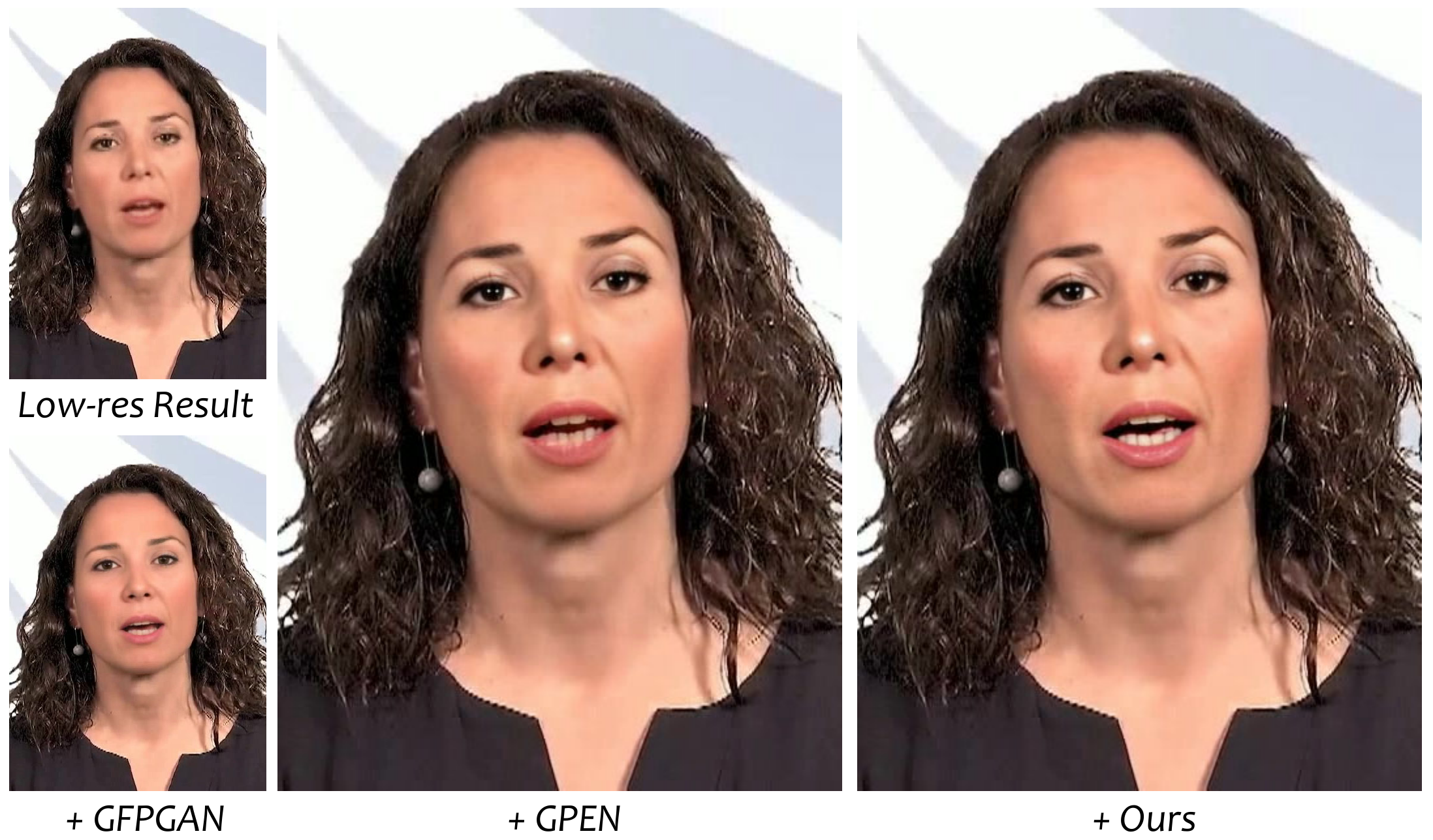}
    \vspace{-1em}
    \caption{Comparison between different face restoration networks on the results, including GFPGAN~\cite{gfpgan}, GPEN~\cite{gpen}, and our hybrid method. Note that, GFPGAN changes identity a lot. Natural face \copyright~\emph{ONU Brasil}~(CC BY).}
    \vspace{-1.5em}
    \label{fig:sr}
\end{figure}

The result from ${L}$-Net is still unperfect since it is hard to train the model on high-resolution talking-head datasets. On the one hand, there is no public available large-scale high-resolution talking-head dataset. 
% To enhance the quality of the final video, we propose an identity-aware enhancement network~($E$-Net). 
On the other hand, if we directly apply the GAN-prior based face restoration networks~\cite{gfpgan, gpen} as the post-processing tools to improve the results, the results might not be perfect in terms of identity changes~\cite{gfpgan} and blurry teeth and face~\cite{gpen} as shown in Figure~\ref{fig:sr}. 

To this end, we propose an identity-aware enhancement network inspired by recent image generation networks~\cite{stylegan2, eg3d}. In detail, to acquire the high-resolution talking-head dataset and aligned domain for up-sampling, we enhance the low-resolution dataset firstly using a GAN prior-based face restoration network~\cite{gpen}. However, there is a domain gap between the enhanced high-resolution dataset during training and the blurry output of $D$-Net during testing. Then, to avoid this gap, we produce the low-resolution input of $E$-Net by feeding the enhanced frame and its corresponding audio to the $L$-Net. Ideally, $L$-Net should produce the same lip motions as the original frame using the conditional audio. Thus, we can use the high-resolution input as supervision directly. As for the architecture, we learn two style-based blocks~\cite{stylegan2} to up-sample the results four times and we design a ResBlock-based encoder ${E}_i$-Net to generate the identity-aware global modulation in each style block.
% to acquire the high-resolution talking-head dataset, we seek help from the GAN prior-based face restoration network~\cite{GPEN} to enhance the LRS2 dataset for training $L$-Net. However, we will get flicker results if we train the $E$-Net on the enhanced datasets since the unstable quality of image-based face restoration network. 
% On the other hand, if
% Thus, we design a progressive training strategy for identity-aware face upsampling.  In training, we resize the enhanced images to low-resolution network and feed the target frame and the corresponding audio to the network. use the pre-trained ${L}$-Net as the lip-generator, and train the ${E}$-Net by feeding the audio and its corresponding face as input. Ideally, ${L}$-Net will produce the same lip shape as the input itself for a paired learning. As for the structure of ${E}-$Net, which contains two style-based blocks ${E}_u-$Net and an reference encoder ${E}_i-$Net for global modulation. 

\subsection{Post-processing}
\label{sec:post}
We also remove several artifacts when pasting back to the original video, including the artifacts of teeth generation and the synthesizing bounding box from the $L$-Net. Synthesizing the photo-realistic teeth for the face video is surprisingly hard~\cite{obama}. Unlike previous approach which uses the teeth proxy~\cite{obama}, we seek help from the pre-trained face restoration network~\cite{gfpgan} for teeth enhancement through face parsing~\cite{face_parser}. As for the face bounding box caused by $L$-Net, we segment~\cite{face_parser} the produced face and paste back to the original video using the Multi-band Laplacian Pyramids Blending~\cite{blend}.

% To reduce the artifacts of face inpainting bounding box, we use a pre-trained face parser network~\cite{face_parser} to segment the generated face and paste it to back to the video using Multi-band Laplacian Pyramids Blending~\cite{blend}. We enhance the teeth the lip region using StyleGAN-based face enhancement network~\cite{gfpgan}.

% \subsubsection{Face Enhancement}

% \subsubsection{Mouth Refinement}
% \xiaodong{\\
% Add a figure: \\
% 1. WAV2LIP+ENHANCE v.s. OURS + ENHANCE \\
% 2. wav2lip method will show noticeable artifacts.
% }
% \input{figs_tex/fig_enhancement}

% \xiaodong{\\
% Add a figure: \\
% 1. comparison between different enhancement methods on face~(original, GFPGANv1.3, GPEN, ours, ground truth) [half-face and full face] \\
% 2. GPEN show the too sharp results and GFPGANv1.3 changes ID. \\
% 3. our method show a more natural results with the mouth from GFPGAN
% }
% \begin{figure}
%     \centering
%     \includegraphics[width=0.9\columnwidth]{figs_pdf/sr_vrt.pdf}
%     \vspace{-1em}
%     \caption{Comparison between different face restoration networks on the results, including GFPGAN~\cite{gfpgan}, GPEN~\cite{gpen}, and our hybrid method. Note that, GFPGAN changes identity a lot.}
%     \vspace{-2em}
%     \label{fig:sr}
% \end{figure}

\begin{figure*}[t]
\begin{center}
\centerline{\includegraphics[width=1\linewidth]{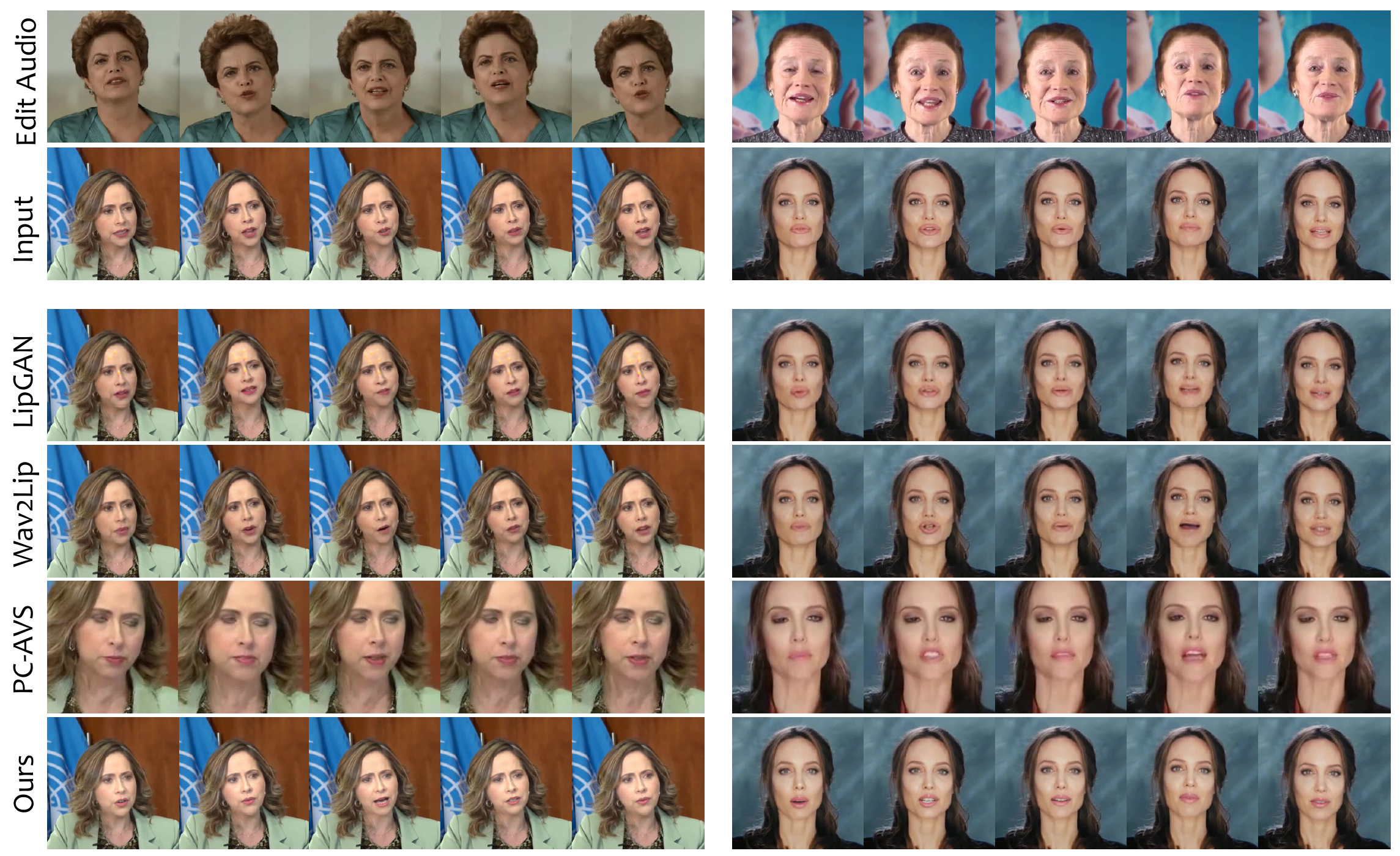}}
\vspace{-2.5mm}
\caption{Qualitative comparison with LipGAN~\cite{lipgan}, Wav2Lip~\cite{wav2lip}, and PC-AVS~\cite{pc-avs}. Above two rows show the edit audio and the input video frames, respectively. Note that, to visualize the input audio, we use the audio's corresponding face to show their mouth shapes.  Natural face \copyright~\emph{ONU Brasil}~(CC BY).}
\setlength{\abovecaptionskip}{-0.1cm}
\label{fig:compare}
\end{center}
\vspace{-3mm}
\end{figure*}
\begin{table*}[!htb]
\centering
\caption{Quantitative results on LRS2 and HDTF datasets.}
\setlength{\abovecaptionskip}{0.1cm}
\vspace{-1em}
\label{tab:quantitative}
\begin{tabular}{l |cccc c cccc}
\toprule
\multirow{3}{*}{}  & \multicolumn{4}{c}{LRS2 Dataset} & & \multicolumn{4}{c}{HDTF Dataset} \\
\cline{2-5} \cline{7-10}
& \multicolumn{2}{c}{Visual Quality} & \multicolumn{2}{c}{Lip-Sync} &  & \multicolumn{2}{c}{Visual Quality} & \multicolumn{2}{c}{Lip-Sync} \\
 & FID$\downarrow$ & CPBD$\uparrow$ & LSE-D$\downarrow$ & LSE-C$\uparrow$ &  & FID$\downarrow$ & CPBD$\uparrow$ & LSE-D$\downarrow$ & LSE-C$\uparrow$ \\
\hline
LipGAN~\cite{lipgan} &       5.168 & 0.2615 & 9.609 & 3.062 & & 7.684 & 0.2754  & 9.943 & 4.052 \\
Wav2Lip w/o GAN~\cite{wav2lip} &     5.069 & 0.2607 & 7.116 & 6.889 & & 7.358 & 0.2764 & \textbf{8.689} & \textbf{5.427} \\
Wav2Lip~\cite{wav2lip} &  \textbf{3.911} & 0.2714 & 7.191 & 6.870 & & 5.632 & 0.2763 & 8.895  & 5.228 \\
%PC-AVS~(One-Shot)~\cite{pc-avs}  &      15.678& 0.2276 & 7.086 & 6.878 & &-&-&-&- \\
PC-AVS~\cite{pc-avs} &       12.800& 0.2085 & 7.666 & 5.974 & &-&-&-&- \\
% Wav2Lip~(Ours run) &         4.966 & 0.2560 & 7.422 & 6.214 & & 23.128 & 0.2725  & 9.818  & 4.177 \\
Ours &                       5.193 & \textbf{0.2809} & \textbf{6.519} & \textbf{7.089} & & \textbf{4.504} & \textbf{0.2903} & 9.359  & 4.518 \\
\bottomrule
\end{tabular}
\vspace{-1em}
\end{table*}
\section{Training}
Our framework is implemented using Pytorch~\cite{pytorch}, and we train each module individually. After training, the whole framework can be tested in a sequence without manual intervention. 
% Move to Supp!
% All the optimizers are Adam. The learning rates of $D$-Net, $L$-Net and $E$-Net are $1e^{-4}$, $ 1e^{-4}$ and $ 1e^{-5}$, respectively. 
Below, we give the dataset and training details of each module.
More details can be found in supplementary material.

\subsection{Training for each module}
\subsubsection{$D$-Net}

% \subsection{Implementation Details}
% \subsection{Experimental Settings}
% \subsubsection{Datasets}
% % The video self-decomposition module is trained on VoxCeleb~\cite{voxceleb} dataset including 22496 talking head videos. . 
% We train the lip synthesis module on LRS2~\cite{lrs2} dataset which consists of 160p videos from BBC programs. We pre-process LRS2 dataset by face detection as in \cite{wav2lip}. HDTF~\cite{hdtf} is a new high-resolution talking-face video dataset containing 720p or 1080p talking head videos from YouTube. We evaluate the proposed method on both low-resolution dataset~(LRS2) and high-resolution dataset~(HDTF). Following the unpaired evaluating setting described in \cite{wav2lip}, we take a video clip and an audio clip from another different video as the inputs. As for the evaluation, we created $14k$ and 100 twenty-second audio-video pairs for LRS2 and HDTF dataset evaluation respectively.

% \subsubsection{Implementation Details}
To perform semantic-guided expression reenactment, we train our network on the VoxCeleb~\cite{voxceleb} dataset with the pose and expression from \cite{face3d}. This dataset contains 22496 talking head videos with diverse identities and head poses. We resize the input frames to 256$\times$256 and train the network on the cropped faces similar to \cite{fomm}. 
% We train $D$-Net on the VoxCeleb~\cite{voxceleb} dataset with the pose and expression from \cite{face3d}. This dataset contains 22496 talking head videos with diverse identities and head poses. We resize the input frames to 256$\times$256 and train the network on the cropped faces similar to \cite{fomm} in 400k iterations using a progressive training setting.
% Move to Supp!
% For training, we randomly choose two frames from the original video as the source and target image pairs, and we reconstruct the target frame using the source image and coefficients of the target frame. 
We train the network in 400k iterations using a progressive training setting.
As for the loss function, we calculate the pixel-wise differences between the predicted image and the ground truth using perception loss~\cite{zhang2018lpips} and ~gram matrix loss~\cite{gatys2016image}.

\subsubsection{$L$-Net}
We train the $L$-Net on the LRS2~\cite{lrs2} dataset. This lip-reading dataset contains large-scale 160p videos from BBC programs. We pre-process the dataset using face detection~\cite{fan} and resize the input image to $96\times96$ following the previous method~\cite{wav2lip}. 
% Move to Supp!
% The audio features are mel-specrograms conducted from 16kHz audio with FFT window size 800 and hop size 200. The $L$-Net is trained in 400k iterations. We use the continuous five frames of the original video and their corresponding windowed audio features as input, the random frames from the same video will be considered as the references in training.
We train the $L$-Net using perceptual loss and lip-sync discriminator for visual quality and audio-visual synchronization~\cite{wav2lip}, respectively.

\subsubsection{$E$-Net}
The training process of $E$-Net is based on $L$-Net. We enhance the LRS2 dataset in advance to get a high-resolution dataset, and train the $E$-Net in 300k iterations. 
% Move to Supp!
% To perform down-sampling, we use a hybrid method of differentiable JPEG and bilinear down-sampling. 
As for the loss function, $E$-Net is trained on the hybrid losses of perceptual loss~\cite{johnson2016perceptual_loss}, pixel-wise $L_1$ loss, adversarial loss~\cite{pix2pix}, lip-sync discriminator~\cite{wav2lip} and identity-loss using a pre-trained face recognition network~\cite{deng2019arcface}.

\subsection{Evaluation}

We evaluate the proposed method in terms of visual quality and lip-synchronization. As for the visual quality, since the ground-truth talking video is unavailable, we choose Fr$\rm\bf\acute{e}$chet inception distance (FID)~\cite{heusel2017fid} and cumulative probability blur detection (CPBD)~\cite{narvekar2009no} to evaluate the visual quality of generated videos. A lower FID score means that the generated images are closer to the dataset distribution. The CPBD reflects the sharpness of the results. Different from~\cite{wav2lip}, we compute visual quality metrics on the full frames of the video instead of cropped faces since we focus on the quality of the whole video. 
% For the lip-synchronization metrics, Landmarks Distance around the mouths~(LMD) cannot be adopted since the lack of ground-truth videos.
We choose the LSE-C and LSE-D~\cite{wav2lip} to evaluate the quality of lip synchronization. As for the dataset choices, we evaluate our framework on both low-resolution dataset~(LRS2) and high-resolution dataset~(HDTF). 
% LRS2 dataset contains 160p videos from BBC programs. HDTF~\cite{hdtf} is a new high-resolution talking-face video dataset containing 720p or 1080p videos from YouTube. 
HDTF dataset contains 720p or 1080p videos from YouTube.
Following the unpaired evaluating settings as described in \cite{wav2lip}, we take a video and an audio clip from the other different video to synthesize the results. We create $14k$ and 100 twenty-second audio-video pairs for LRS2 and HDTF dataset evaluation respectively.

\section{Results}
\label{sec:result}
\subsection{Comparison with state-of-the-art Methods}
% \subsubsection{Comparison Methods}
We compare our method with three state-of-the-art methods under the same settings, including LipGAN~\cite{lipgan}, Wav2Lip~\cite{wav2lip} and PC-AVS~\cite{pc-avs}. 
LipGAN and Wav2Lip share the similar network structures. Differently, Wav2Lip uses a pre-trained lip-sync discriminator as the lip-expert, yet a better lip-sync performance. PC-AVS is originally proposed for one-shot pose-controllable talking-head generation. We use the identity code of each original video frame to replace the original single image face animation settings. We compare the proposed method with these methods using their open-sourced codes.

% \subsubsection{Quantitave Comparison}

% \xiaodong{\\
% 1. $L$-Net lama baseline l1 \\
% 2. $L$-Net LamaTrans l1 \\
% 3. $L$-Net + $E$-Net: lamasrv2 + gfpgan~(for lip enhancement) + face parser~(for parser) \\
% 4. $L$-Net + $E$-Net + $D$-Net : based on 3 + pirender~(video decomposition) \\
% }

% \begin{table*}[!htb]
% \centering
% \caption{Quantitative results on LRS2 and HDTF datasets.}
% \label{tab:quantitative_}
% \begin{tabular}{c ccccc c ccccc}
% \toprule
%   & \multicolumn{5}{c}{LRS2} & & \multicolumn{5}{c}{HDTF} \\
% \cline{2-6} \cline{8-12}
% % & \rotatebox{90}{LipGAN} & \rotatebox{90}{Wav2Lip} & \rotatebox{90}{Wav2Lip+GAN} & \rotatebox{90}{PC-AVS} & \rotatebox{90}{Ours} & & \rotatebox{90}{LipGAN} & \rotatebox{90}{Wav2Lip} & \rotatebox{90}{Wav2Lip+GAN} & \rotatebox{90}{Ours} & \rotatebox{90}{Ours+Enhance} & \rotatebox{90}{Full Pipeline} \\
% & LipGAN & Wav2Lip & Wav2Lip+GAN & PC-AVS & Ours & & LipGAN & Wav2Lip & Wav2Lip+GAN & Ours+Enh & Ours+Full \\
% \cline{1-12}
% % \midrule
% FID   \\
% CPBD \\
% LSE-D \\
% LSE-C \\
% \bottomrule
% \end{tabular}
% \end{table*}

% \subsubsection{Qualitative Comparison}
As shown in Table~\ref{tab:quantitative}, the proposed method achieves much better visual qualities according to CPBD and FID. Since the LRS2 dataset is low-resolution and our method produces high-resolution results, the FID of Wav2Lip on the LRS2 dataset is better. As for the accuracy of lip-sync, our method still gets much better and comparable performance on these two datasets. We also show some examples in Figure~\ref{fig:compare} to perform the visual comparison. From this figure, our method produces high-quality results with more accurate lip-sync than previous methods. Since visual dubbing is a video editing task, we highly recommend the reader to compare our methods with others refer to the accompanying video.

% \subsubsection{User Study}
% The lip synchronization quality is necessary to be evaluated by the human. 
For the comparison of the lip-sync quality, human evaluation is required. We perform a user study to further evaluate the performance of the proposed method. In the user study, we generate ten talking videos with different audio and video sources of our method and two state-of-the-art methods~(LipGAN and Wav2Lip) on the HDTF dataset. We let the users show their opinions about each video in terms of the visual and lip-sync qualities. We set five different scores~(larger is better, ranging from 1 to 5) for each option. Our form is sent to 51 people in total, getting 510 opinions. As shown in Table~\ref{tab:user}, most users prefer to give higher scores to our method with respect to the visual and lip-sync quality.

\begin{table}[h]
\centering
% \vspace{-1em}
\caption{User Study.}
\vspace{-1em}
\label{tab:user}
\begin{tabular}{lcccc}
\toprule
% \multirow{2}{*}{Method} & \multicolumn{4}{c}{LRS2} \\
 Method & Visual Quality$\uparrow$ & Lip-Sync  Quality$\uparrow$ \\
\hline
LipGAN &    2.867 & 3.058    \\
Wav2Lip &   3.173 & 3.398   \\
Ours &      \textbf{4.171} & \textbf{4.100}  \\ 
\bottomrule 
\end{tabular}
\end{table}

% \xiaodong{add a table for user study.}

\subsection{Ablation Study}
We mainly ablate three major components of our framework in Table~\ref{tab:ablation}. 
% \sout{Inspired by recently general image inpainting network~\cite{lama}, we design a baseline model and modify it to suitable our task by the additional image encoder and audio conditional.}
The first component is the cross-attention between two image encoders. 
$L$-Net w/o cross-attention in Table~\ref{tab:ablation} means channel-wisely concatenating the features from the source and reference frames.
We find cross-attention is helpful in terms of the lip-sync quality since it can capture the long-range dependencies. Besides the gains in numerical metrics, we also find it brings more vivid results~(e.g, larger mouth). We then show the results of adding the $E$-Net in our framework. As we expected, the identity-aware face enhancement will hugely improve the visual quality. However, the additional artifacts will also influence the lip-sync quality. Finally, by using $D$-Net to stabilize reference frames, our framework generates better video in terms of visual and lip-sync quality.

\begin{table}[h]
\centering
\caption{Major Ablation Studies on HDTF Dataset.}
\vspace{-1em}
\label{tab:ablation}
\resizebox{\columnwidth}{!}{
\begin{tabular}{l|cc|cc}
\toprule
% \multirow{2}{*}{Method} & \multicolumn{4}{c}{LRS2} \\
& \multicolumn{2}{c|}{Visual Quality} &  \multicolumn{2}{c}{Lip-Sync Quality} \\
& FID$\downarrow$ & CPBD$\uparrow$ & LSE-D$\downarrow$ & LSE-C$\uparrow$ \\
\hline
% Wav2Lip~(Ours run) &                 7.722 & 0.2770 & 9.430  & \textbf{4.577} \\
% $L_{basic}$-Net~(Ours) &              6.471 &  &   &  \\
$L$-Net w/o cross-att.  &     5.951 & 0.2743 & 9.788  & 4.164 \\
% $L$-Net w/ FFC &     - & - & -  & - \\
$L$-Net   &  6.471 & 0.2755 & 9.578  & 4.382 \\ 
$L$-Net + $E$-Net &           \textbf{3.334} & 0.2873 & 10.171 & 3.764 \\ 
$L$-Net + $E$-Net + $D$-Net&  4.504 & \textbf{0.2903} & \textbf{9.359}  & \textbf{4.518}  \\ 
\bottomrule 
\end{tabular}
}
\end{table}

\subsection{Extensions to Emotional Talking Video}
\begin{figure}[t]
    \centering
    \includegraphics[width=\columnwidth]{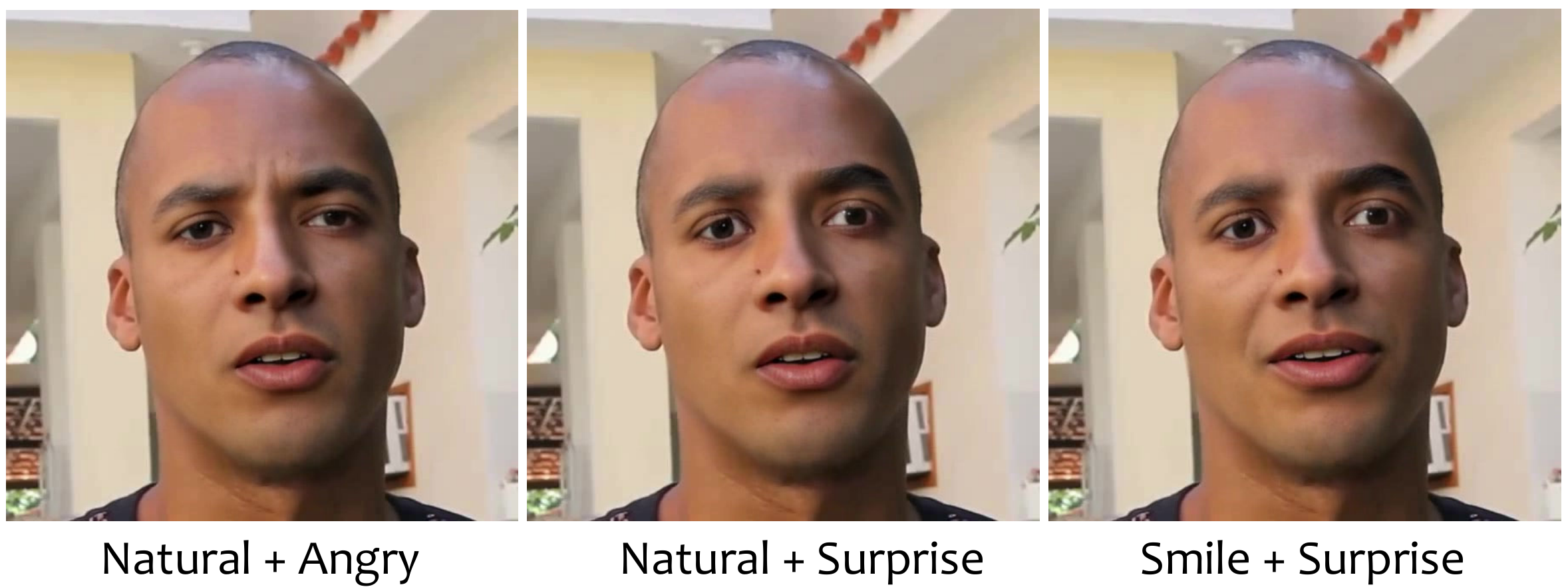}
    % \vspace{-1em}
    \setlength{\abovecaptionskip}{0.1cm}   % distance between the caption and Figure
    \caption{More emotional results using \cite{ganimation}. Natural face \copyright~\emph{ONU Brasil}~(CC BY).}
    % \vspace{-1em}
    \label{fig:emo}
\end{figure}

We have already shown that the proposed method can be used for emotional talking-head video editing in Figure~\ref{fig:teaser}. Since our method only modifies the lower-half face, we also get inspiration from the facial action unit system~\cite{facs} and edit the upper face of the images using \cite{ganimation}, causing different combinations as shown in Figure~\ref{fig:emo}. 
% since their method is an image based method, we reduce the artifacts and get more photo-realistic results of the upper face using \cite{gpen}.
% \xiaodong{add a figure to different emotion by GANimation.}

\begin{figure}[t]
    \centering
    \includegraphics[width=\columnwidth]{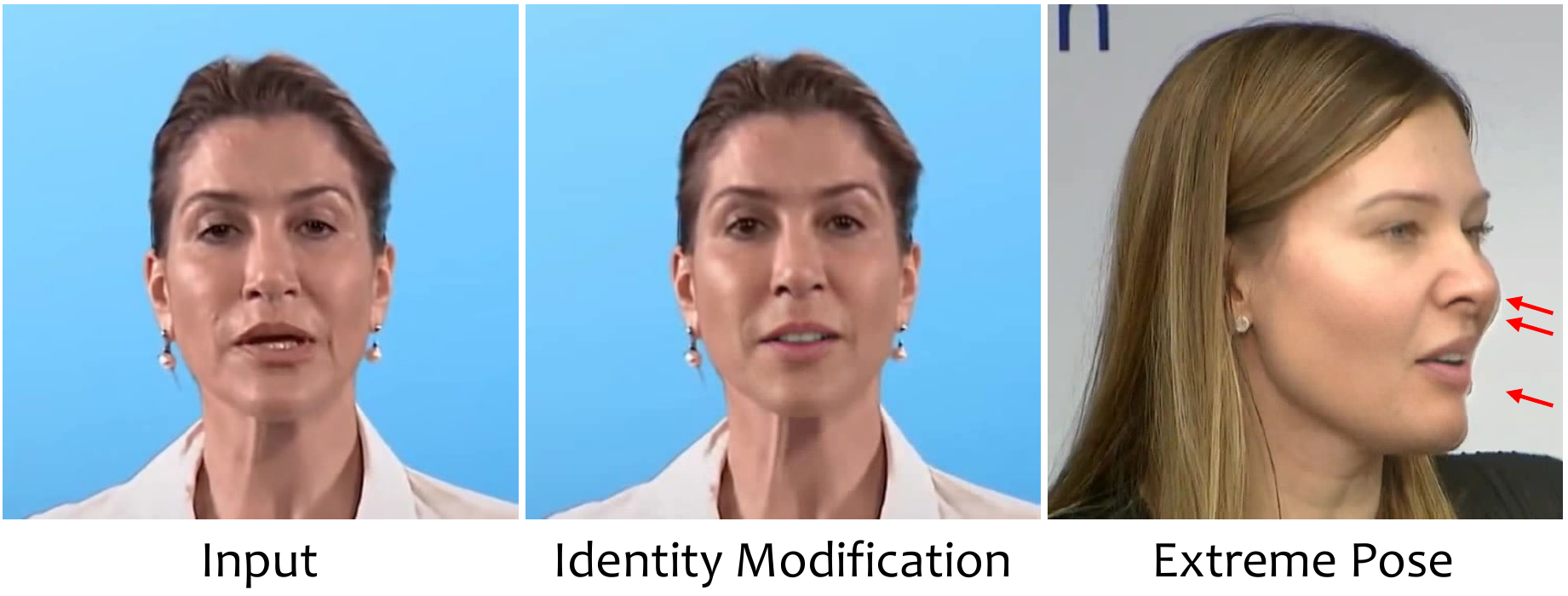}
    % \vspace{-1em}
    \setlength{\abovecaptionskip}{0.1cm}
    \caption{Failed cases on identity and extreme poses. Natural faces \copyright~\emph{ONU Brasil} and \copyright~\emph{European Central Bank}~(CC BY).}
    % \vspace{-0.5em}
    % \vspace{-1em}
    \setlength{\belowcaptionskip}{-0.1cm}
    \label{fig:failed_cases}
\end{figure}

\subsection{Limitation}
Although the proposed method can work for the videos in the wild, it still contains some noticeable artifacts in some cases. As shown in Figure~\ref{fig:failed_cases}, one noticeable difference of the proposed framework will cause a slightly identity change from the original video due to the dense warping of $D$-Net. However, it is only one module of our method and we will replace it with another face reenactment network~\cite{face_vid2vid} or 3D-based face reenactment method~\cite{dvp} directly. Our method also shows some artifacts in some extreme poses as shown in Figure~\ref{fig:failed_cases}. Since our method edits the video in a frame-by-frame fashion, the results may show some small temporal jittering and flashing.

\section{Conclusion}
We present a generic system for audio-based talking-head video editing by removing the lip motion first and then performing editing. As demonstrated, our framework can work on in-the-wild videos without fine-tuning and produce high-quality results using the audio as the condition. 
% The proposed method edits the lower-half face of the video according to audio guidance, resulting a high-quality and accurate lip synchronization video for in the wild videos.
Besides, our system has the potential on the emotional talking-head generation for the lower-half face of the video. We will explore in the future to support more emotions and connect the source audio and contexts to the emotions. 

\textbf{Ethical Considerations.} Since our system can edit the talking content of the video in the wild, we also consider the misuse of the proposed method.
% To prevent the misuse of the synthesized video, we will add both robust video and audio watermark to the produced video, and develop the tools to identify the trustworthiness. On the other hand, we hope our method can also help the research in the DeepFake detection. 
We will add both robust video and audio watermark to the produced video, and develop the tools to identify the trustworthiness. On the other hand, we hope our method can also help the research in the DeepFake detection.

\begin{acks} 
This work was supported in part by the National Key Research and Development Program of China under Grant 2018AAA0103202, in part by the National Natural Science Foundation of China under Grant 61922066, 61876142 and 62036007, in part by the Technology Innovation Leading Program of Shaanxi under Grant 2022QFY01-15, in part by Open Research Projects of Zhejiang Lab under Grant 2021KG0AB01.
\end{acks}

% \clearpage

%%
%% The next two lines define the bibliography style to be used, and
%% the bibliography file.
\bibliographystyle{ACM-Reference-Format}
\bibliography{reference.bib}

\clearpage
\appendix
\section{Study on Different Expression Templates}
We use a hand-crafted neutral expression template for all the experiments in our main paper. Here, we provide a study on how different expression templates influence the performance of $D$-Net. To achieve this goal, 
%It is worth noting that only the lower-half normalized face is sent to $L$-Net, so we choose neutral and smile expressions for this experiment since they have distinct lower-face emotions. 
we interpolate the expression templates between the neutral and smile expression coefficients and evaluate the results on 10 videos from HDTF~\cite{hdtf} dataset. In Table \ref{tab:templates}, the lip-sync metrics show minor changes when we edit the expression templates, indicating that $D$-Net is robust to the expression template.

\begin{table}[h]
\centering
% \vspace{-1em}
\caption{Lip-Sync metrics on 10 videos from HDTF dataset using different expression templates.}
\vspace{-1em}
\label{tab:templates}
\begin{tabular}{lcc}
\toprule
\makecell[c]{Interpolation Ratio \\ from Neutral to Smile Expression} & LSE-D $\downarrow$ & LSE-C $\uparrow$ \\
\hline
 0~(Neutral Template) & 9.027 & 4.829 \\
 0.2         & 9.058 & 4.798 \\
 0.4         & 9.092 & 4.768 \\ 
 0.6         & 9.084 & 4.781 \\
 0.8         & 9.025 & 4.834 \\
 1~(Smile Template)  & 8.924 & 4.929 \\
\bottomrule 
\end{tabular}
\end{table}
\section{Analysis of the Trade-off Between Identity Preservation and Expression Animation}

Our method utilizes the information leaks between models for expression editing. Interestingly,
% $D$-Net is specially designed to prevent information leaks. However, the generated videotiny lip animation may still appear since there are unlimited expressions in the source video, resulting in imperfect prediction. 
we find there is a trade-off between the identity preservation and expression editing when using $D$-Net. In detail, the expression normalization can be done in both one-shot face reenactment and video to video settings for better expression animation and identity preservation, respectively. 
In one-shot setting, the reenacted video is reconstructed by warping the first frame of the whole video using the original pose coefficients and template expression coefficients. In this setting, $D$-Net can generate more stable lip animation, yet a better lip-sync performance. However, since $D$-Net uses the dense flow for warping, there is a little identity modification.
% To modify the pose from the first frame to the current frame, the pose coefficients are extracted from the current frame and the expression coefficients are extracted from the expression template.
In video-to-video setting, we do not fix the reference frame to be warped, allowing $D$-Net to change the expression of each frame, which helps identity preservation but cause a little unstable on lip movement. We choose this setting as the default choice in our method.

% For better identity preservation, we choose the many-shot setting in our method.

\section{Implementation Details}

\subsection{Implementation Details of D-Net}
% \textbf{Model Architecture.} 
\subsubsection{Model Architecture} 
% MappingNet. 
% WarppingNet. 
% EditingNet.
The architecture design of $D$-Net is similar to PIRenderer~\cite{pirender}, which consists of three sub-networks for coefficient mapping, feature warping and refinement. The driven 3DMM coefficients will be translated to the latent codes $z$ through the mapping network and then injected into the feature warping and refinement networks. In detail, the mapping network contains four 1D convolution layers and applies the Leaky-ReLU as activation function to calculate the global feature. The architecture of the warping network and editing network is an encoder-decode-based network with skip connections. 
The warping network does downsampling five times and upsampling three times, and then generates flow fields that are a quarter of the original size. The editing network contains three stages to learn multi-scale features. 
We use the CONV-SpertralNorm-LeakyReLU block as the up-sampling and down-sampling layers. The AdaIN~\cite{huang2017adain} blocks are applied after each convolution layer to inject the motion information.

\subsubsection{Loss Functions} 
% Warpping Network: perceptual loss
% Editing Network: perceptual loss, gram loss
For the warping network, we calculate the perceptual loss~\cite{johnson2016perceptual_loss} between the warped image $I_{D_w}$ and ground truth:
\begin{equation}
\mathcal{L}_{D_w} = \mathcal{L}_{perceptual} = \sum_l \vert\vert\ f_{vgg}^l(I_{gt}) - f_{vgg}^l(I_{D_w})\vert\vert_2,
\end{equation}
where $f_{vgg}$ is the pre-trained $VGG$-19 network~\cite{simonyan2014vgg} and $l$ is the layer of the feature map.

For the editing network, we calculate the perceptual loss and gram matrix style loss~\cite{gatys2016image} between the generated image $I_D$ of the whole $D$-Net and ground truth:
\begin{equation}
\mathcal{L}_{c} = \sum_l \vert\vert\ f_{vgg}^l(I_{gt}) - f_{vgg}^l(I_D)\vert\vert_2,
\end{equation}

\begin{equation}
\mathcal{L}_{s} = \sum_l \vert\vert\ G(f_{vgg}^l(I_{gt})) - G(f_{vgg}^l(I_{D}))\vert\vert_2,
\end{equation}
where G is the gram matrix constructed from activation map.
The full optimization objective of editing network is:
\begin{equation}
\mathcal{L}_{D_e} = \lambda_{c} \mathcal{L}_{c} + \lambda_{s} \mathcal{L}_{s},
\end{equation}
\noindent where $\lambda_{c}=1$ and $\lambda_{s}=250$.

\subsubsection{Training and Inference Details}
% Pre-train the Warpping Network and then train the Editing network.
% Training: self-reconstruction; Testing: 
We train $D$-Net on the VoxCeleb~\cite{voxceleb} dataset. We first pre-train the mapping network and the warping network for 200k iterations, and then train the whole network for another 200k iterations. We use Adam~\cite{adam} optimizer and the learning rate is $1e^{-4}$. 
% During the training phase, the network is doing a self-reconstruction task as described in our main paper.
% Move from main paper!
During the training phase, the network is doing a self-reconstruction task. We randomly choose two frames from the original video as the source and target image pairs, and we reconstruct the target frame using the source image and coefficients of the target frame. 
During the testing phase, we use the pose coefficients of the source image and expression coefficients of pre-defined template to normalize the lip shape while maintaining the original pose of the video.

\subsection{Implementation Details of L-Net}

\begin{figure*}
    \centering
    \includegraphics[width=1\linewidth]{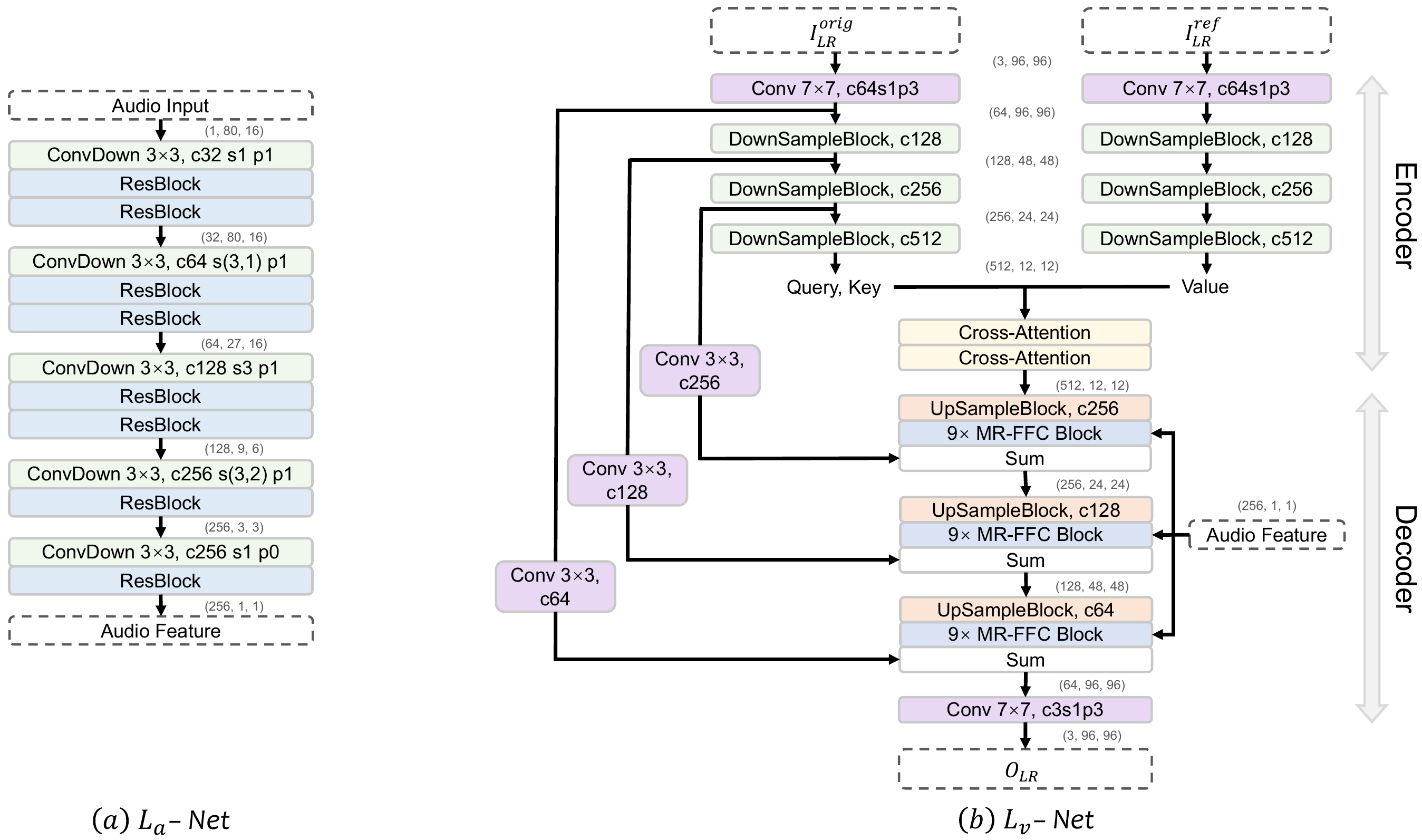}
    % \vspace{-2.5em}
    \caption{The architecture of the $L$-Net.}
    % \vspace{-4mm}
    \label{fig:supp_LNet}
\end{figure*}
\begin{figure*}
    \centering
    \includegraphics[width=1\linewidth]{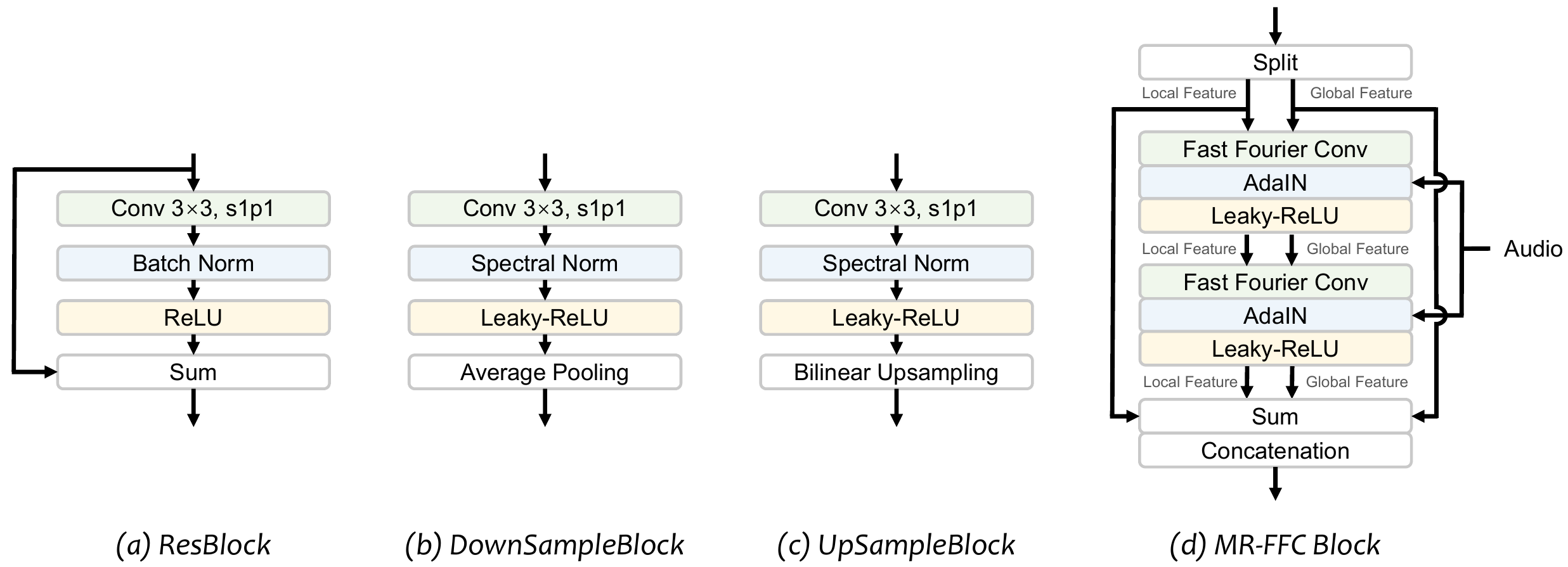}
    % \vspace{-2mm}
    \caption{The components used in the $L$-Net. (a)ResBlock, (b)DownSampleBlock, (c)UpSampleBlock, and (d)LaMa-AdaIN Block.}
    % \vspace{-2mm}
    \label{fig:supp_blocks_LNet}
\end{figure*}
\begin{figure*}
    \centering
    \includegraphics[width=0.8\linewidth]{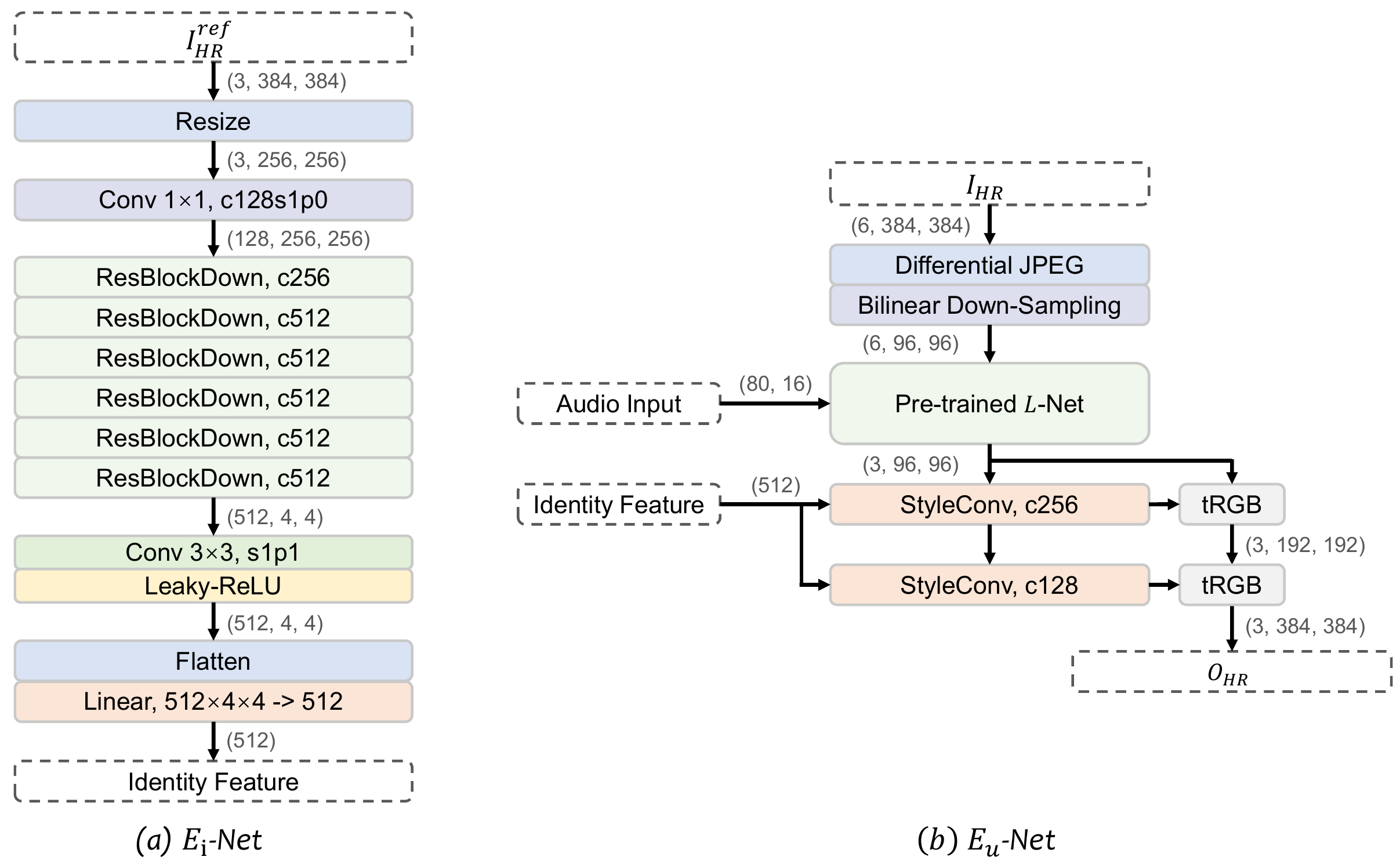}
    % \vspace{-2.5em}
    \caption{The architecture of the $E$-Net.}
    % \vspace{-4mm}
    \label{fig:supp_ENet}
\end{figure*}

\subsubsection{Model Architecture}
% Encoder and Decoder; Cross-attn.; Lama; AdaIN
The $L$-Net consists of an audio encoder network $L_a$-Net and a visual encoder-decoder-based network $L_v$-Net, as shown in Figure~\ref{fig:supp_LNet}.
The audio encoder $L_a$-Net, which consists of several ResBlock-based down-sampling layers, are used to extract the high-level audio features. The encoder of $L_v$-Net is made up of three down-sampling layers which consist of 2D convolution, batch normalization and Leaky-ReLU activation function. Applying two separate encoder networks, the half-masked original frame $I_{LR}^{orig}$ and randomly selected reference frame $I_{LR}^{ref}$ from the same video are encoded to features $F_{orig}$ and $F_{ref}$, respectively, and then, these features are fused to $F_v$ using two cross-attention blocks~\cite{transformer} for long-range dependencies and avoid the local information leak. For the cross-attention block, we use $F_{orig}$ to generate query $Q$ and key $K$, and then use $F_{ref}$ to generate value $V$. We calculate the attention score between $Q$ and $K$ and the weighted sum of $V$ to obtain the $F_v$. 
The decoder of $L_v$-Net is made up of three up-sampling layers, each of which consist of a convolution-up block, nine Modulated Res-FFC~(MR-FFC) blocks and skip connection from visual encoder. More details about the Fast Fourier Convolution~(FFC) can be found in \cite{ffc}. All the blocks used in $L$-Net are shown in Figure~\ref{fig:supp_blocks_LNet}.

% \noindent \textbf{Loss Functions.}
\subsubsection{Loss Functions} 
% l1 + perc + sync
For the visual quality, we calculate the pixel-wise $L_1$ loss in RGB space and perceptual loss in feature space between the generated results $O_{LR}$ of $L$-Net and low-resolution ground truth $I_{gt}$:
\begin{equation}
\mathcal{L}_1 = \vert\vert I_{gt}-O_{LR}\vert\vert_1,
\end{equation}

\begin{equation}
\mathcal{L}_{perceptual} = \sum_l \vert\vert\ f_{vgg}^l(I_{gt}) - f_{vgg}^l(O_{LR})\vert\vert_2.
\end{equation}

For the audio-visual synchronization, we use pre-trained SyncNet~\cite{wav2lip, syncnet} as lip-sync discriminator to calculate sync loss between continuous five frames:
\begin{equation}
\mathcal{L}_{sync} = \frac{1}{N} \sum_{i=1}^N -\log(P_{sync}),
\end{equation}
\begin{equation}
P_{sync}=\frac{v\cdot a}{\max{\vert\vert v \vert\vert_2 \cdot \vert\vert a \vert\vert_2}},
\end{equation}
\noindent where $P_{sync}$ indicates the probability that the input audio-video pair is in sync. $v$ and $a$ are video and audio embeddings extracted from pre-trained SyncNet.
The full optimization objective of $L$-Net is:
\begin{equation}
\mathcal{L}_{L} = \lambda_{1} \mathcal{L}_{1} + \lambda_{p} \mathcal{L}_{perceptual} + \lambda_{sync} \mathcal{L}_{sync},
\end{equation}
\noindent where $\lambda_{1}=1$, $\lambda_{p}=1$ and $\lambda_{sync}=0.3$.

\subsubsection{Training and Inference details}
% Training: Random selected frames;
% Inference: Current frames;
We train and inference the $L$-Net following the pipeline of Wav2Lip~\cite{wav2lip}. During the training phase, the input frames $I_{LR} \in \mathbb{R}^{5\times6\times96\times96}$ are made up by five continuous frames with five randomly selected reference from the same video. We also send the corresponding audio window to the network for driving information. 
% Move from main paper!
The audio features are mel-specrograms conducted from 16kHz audio with FFT window size 800 and hop size 200. We train $L$-Net in 400k iterations on the LRS2 dataset using Adam optimizer. The learning rate of $L$-Net is $1e^{-4}$. 
During the testing phase, we use the whole input frames as the reference frame to preserve the pose and background information.

\subsection{Implementation Details of E-Net}
As discussed in the main paper, E-Net is used to upsample the generated videos by the enhanced LRS2 dataset. Below, we give the details of the implementation and training details.

\subsubsection{Model Architecture}
\begin{figure}
    \centering
    \includegraphics[width=\columnwidth]{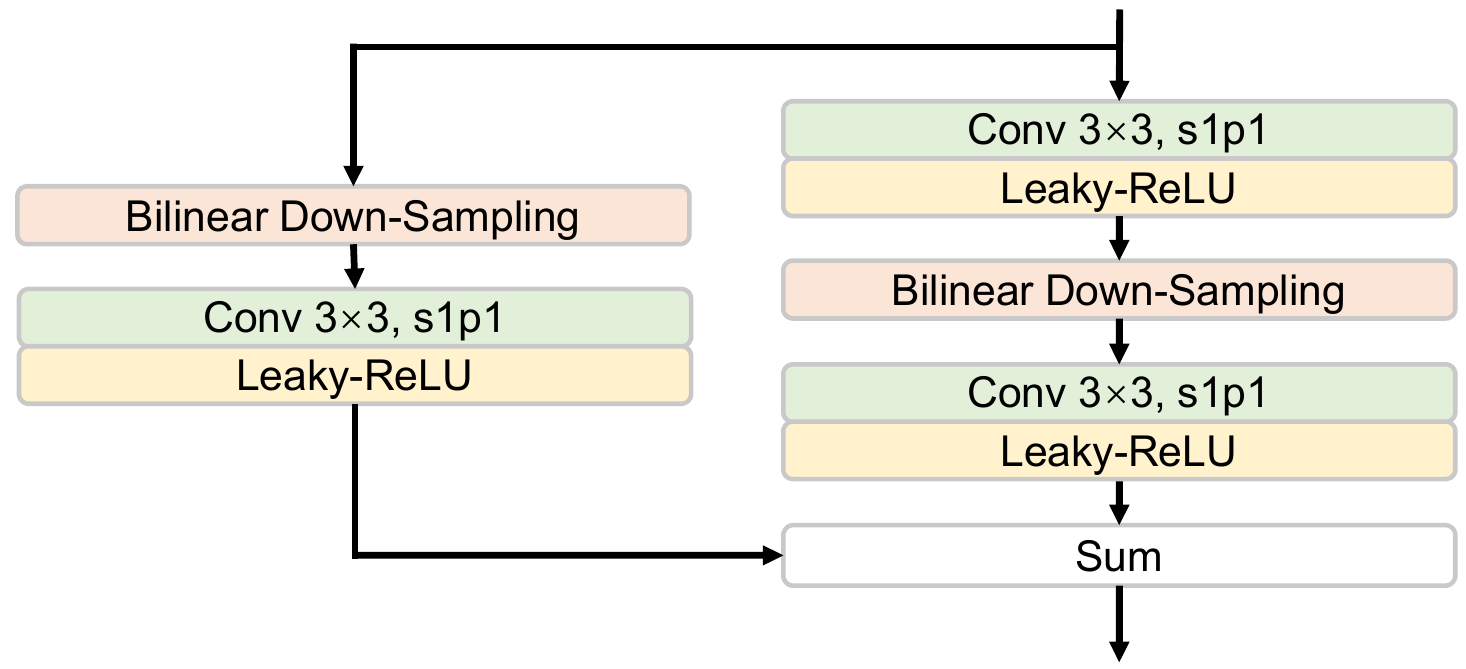}
    % \vspace{-2mm}
    \caption{The architecture of ResBlockDown used in $E_i$-Net.}
    % \vspace{-2mm}
    \label{fig:supp_ResBlockDown_ENet}
\end{figure}

Figure~\ref{fig:supp_ENet} shows the architecture of $E$-Net, which contains an identity encoder $E_i$-Net and a super-resolution module $E_u$-Net. In $E_u$-Net, similar to $L$-Net, we feed the continuous five frames of the video frames $I_{HR}$~(concatenation of masked the lower-half face and itself) and the randomly picked references $I_{HR}^{ref}$ from the same video to the $E_i$-Net. Then, these images will be down-sampled with some augmentations~(including the differential JPEG compression and bi-linear down-sampling) and send to the pre-trained $L$-Net for lip synchronization. 
After that, we get edited frames by the audio to further up-sampling. Inspired by StyleGAN~\cite{stylegan2}, the super-resolution module $E_u$-Net uses the similar blocks to upsample the low-resolution results. Each style-based layer is built by a StyleConv block and a tRGB block to learn the high-resolution results. More details about the StyleConv and tRGB blocks can be found in \cite{stylegan2}.  
% of the $E$-Net is the concatenation of the masked original frame $I_{HR}^{orig}$ and randomly selected reference frame $I_{HR}^{ref}$ from the same video.
%The high-resolution input frame is firstly downsampled to a vector through $E_i$-Net to encode global identity feature and then injected to $E_u$-Net to perform identity-aware enhancement.
In each StyleConv, we also use the features from the identity encoder $E_i$-Net as the modulation for identity preservation. $E_i$-Net is a res-block~\cite{resnet} based encoder which consists of six down-sampling layers and a linear layer. We first resize the high-resolution randomly selected reference frames $I_{HR}^{ref}$ from $384\times384$ to $256\times256$. Then, the down-sampling layers will be used to extract the high-level feature of the $I_{HR}^{ref}$ to a $512$-dimension vector. The detailed architecture of the ResBlock-based down-sampling layer is shown in Figure~\ref{fig:supp_ResBlockDown_ENet}. 

\subsubsection{Loss functions}
We calculate the pixel-wise $L_1$ loss in RGB space and perceptual loss in feature space between the generated results $O_{HR}$ of $E$-Net and high-resolution ground truth $I_{GT}$:
\begin{equation}
\mathcal{L}_1 = \vert\vert I_{GT}-O_{HR}\vert\vert_1,
\end{equation}

\begin{equation}
\mathcal{L}_{perceptual} = \sum_l \vert\vert\ f_{vgg}^l(I_{GT}) - f_{vgg}^l(O_{HR})\vert\vert_2.
\end{equation}

For better identity preservation, we apply the identity loss. Specially, we adopt the pre-trained face recognition network ArcFace~\cite{deng2019arcface} and calculate this loss in feature space similar to perceptual loss:
\begin{equation}
\mathcal{L}_{id} = \vert\vert\ f_{arcface}(I_{GT}) - f_{arcface}(O_{HR})\vert\vert_2.
\end{equation}

To increase the realistic of the generated sample, we also use the adversarial loss:
\begin{equation}
\mathcal{L}_{adv}(G_E,D) = \mathbb{E}_{I_{GT}}[\log D(O_{HR})] + \mathbb{E}_{O_{HR}}[\log (1-D(G_E(O_{HR})))].
\end{equation}

Finally, the full optimization objective of $E$-Net is:
\begin{equation}
\begin{aligned}
(G^*_E, D^*) = \arg\,\min_{G_E}\,\max_D\ & \lambda_{1} L_{1} + \lambda_{p} \mathcal{L}_{perceptual} \\
&+ \lambda_{adv} \mathcal{L}_{adv} + \lambda_{id} \mathcal{L}_{id},
\end{aligned}
\end{equation}
\noindent where $\lambda_{1}=0.2$, $\lambda_{p}=1$, $\lambda_{adv}$=100 and $\lambda_{id}=0.4$.

% \noindent \textbf{Training details.}
\subsubsection{Training details}
To train the $E$-Net, we first perform the face restoration network GPEN~\cite{gpen} to the LRS2 dataset to obtain the high-resolution training dataset. 
% At the training stage of $E$-Net, we do not use $D$-Net and the $L$-Net is pre-trained. 
Then, we use a hybrid data argumentation method of differentiable JPEG and bi-linear down-sampling to get the low-resolution~($96\times96$) input $I_{LR}$ of $L$-Net. Notice that, we do not use the original image as the low-resolution sample since there is still a domain gap to avoid the temporal jitting. Driving by the driven audio, we can get the low-resolution lip-synced result for $E$-Net. Ideally, $L$-Net should produce the same lip motions as the original high-resolution input and we use it as the supervision for upsampling. This network is trained in 300k iterations. We use Adam optimizer and the learning rate is $1e^{-5}$.

\end{document}